\documentclass[manuscript,screen]{acmart}
\usepackage{pdflscape}
\usepackage{booktabs}
\usepackage{longtable}
\usepackage{multirow}
\usepackage{array}
\usepackage[table]{xcolor}
\AtBeginDocument{%
  }

\begin{document}

\title{FLARE: A Systematic, Uncertainty-Aware Framework for Evidence-Based Adoption of Artificial Intelligence in Healthcare}

\author{Jacob Idoko}
\email{jacob.idoko@ucalgary.ca}
\affiliation{%
  \institution{University of Calgary}
  \department{Department of Electrical and Software Engineering}
  \city{Calgary}
  \state{Alberta}
  \country{Canada}
}

\author{Siddhartha Paudel}
\email{siddhartha.paudel1@ucalgary.ca}
\affiliation{%
  \institution{University of Calgary}
  \department{Department of Electrical and Software Engineering}
  \city{Calgary}
  \state{Alberta}
  \country{Canada}
}

\author{Mariana Bento}
\email{mariana.pinheirobent@ucalgary.ca}
\affiliation{%
  \institution{University of Calgary}
  \department{Department of Biomedical Engineering}
  \city{Calgary}
  \state{Alberta}
  \country{Canada}
}

\author{Roberto Souza}
\affiliation{%
  \institution{University of Calgary}
  \department{Department of Electrical and Software Engineering}
  \city{Calgary}
  \state{Alberta}
  \country{Canada}
}

\author{Gouri Ginde}
\email{gouri.ginde@ucalgary.ca}
\affiliation{%
  \institution{University of Calgary}
  \department{Department of Electrical and Software Engineering}
  \city{Calgary}
  \state{Alberta}
  \country{Canada}
}


\begin{abstract}

Artificial intelligence is increasingly being introduced into healthcare workflows, yet most evaluations emphasize model accuracy rather than whether adoption is economically worthwhile in real clinical settings. This study proposes FLARE, a systematic and uncertainty-aware framework for evaluating the financial and operational implications of adopting AI in healthcare. FLARE combines fuzzy logic, time-driven activity-based costing, and return on investment analysis to estimate the cost of clinical service delivery, the cost of AI development and operation, and the economic consequences of workflow integration under uncertainty. The framework was demonstrated through an early health technology assessment case study of AI-assisted large vessel occlusion detection in the CT stroke pathway for acute ischemic stroke.
The case study shows how FLARE can quantify conventional pathway cost, AI-related development and recurring costs, and AI-enabled service savings within a unified activity-based model. Under expected assumptions, the analysis identified a break-even threshold of approximately 3,992 patients per year, with positive first-year return on investment at typical annual stroke volumes of about 5,000 patients. The results further show that economic benefit depends not only on algorithmic performance, but also on patient volume, verification time, infrastructure choices, and workflow design.
FLARE provides a transparent and practical decision-support framework for early-stage evaluation of AI adoption in healthcare. By making uncertainty, resource use, and implementation trade-offs explicit, it helps clinicians, administrators, and policymakers determine when AI deployment is economically viable and where operational changes may improve value.
\end{abstract}




\maketitle

\section{Introduction}\label{sec1}

\paragraph{Motivation.}
The integration of artificial intelligence (AI) into healthcare systems continues to expand, with applications supporting diagnosis, workflow optimization, and clinical decision-making \cite{huang2024artificial} \cite{zhou2021review}. While many AI models now demonstrate strong performance in controlled research settings, the extent to which they translate into meaningful value in real clinical workflows remains unclear \cite{fasterholdt2022value}. Most research stops at reporting performance metrics such as accuracy, offering little insight into whether adopting such AI systems is \emph{economically viable} for healthcare organizations \cite{Bharadwaj2024}. In practice, the integration of AI into clinical workflows introduces new costs related to development and ongoing maintenance \cite{zhang2025economic}.

\paragraph{Problem statement.}
As a result, hospitals and policymakers increasingly require not only evidence of clinical benefit but also a clear understanding of the financial and operational implications of AI adoption. However, existing approaches fail to provide a comprehensive, transparent accounting of the costs and benefits associated with adopting AI solutions throughout their full lifecycle.
This lifecycle perspective is crucial because, as highlighted in the literature, most AI evaluations stop at model accuracy or deployment cost. Addressing this gap requires a framework capable of tracing resource use and time commitments not only during service delivery, but also during the development, validation, deployment, and operational phases of an AI system.

\paragraph{Research objectives.}
The objective of this research is to estimate the ROI of adopting an AI solution across its entire lifecycle, from development and validation to deployment, integration, and sustained use within healthcare workflows.
To achieve this, the research pursues three specific objectives:
(i) to estimate the cost of delivering a clinical service at the activity level under uncertainty;
(ii) to estimate the full cost of an AI solution across development, deployment, and recurring operation; and
(iii) to combine both into ROI, break-even, and sensitivity measures that support early-stage adoption decisions.

To meet these objectives, we propose FLARE (Fuzzy-Logic, Time-Driven Activity-Based ROI), a systematic, uncertainty-aware framework designed to quantify the economic value of adopting AI solutions in healthcare. We built FLARE on TDABC because the literature shows that TDABC is well suited to settings that require granular costing, patient-level resource tracing, and evaluation of care pathways over time \cite{keel2017time}\cite{kaplan2011solve}. By incorporating fuzzy logic, FLARE captures uncertainty in activity durations and resource utilization, and extends these cost estimates into an ROI evaluation that supports evidence-based decision-making for healthcare systems, hospital administrators, and policy stakeholders.

\paragraph{Contributions.}
This paper makes the following contributions:
\begin{itemize}
    \item \textbf{The FLARE framework.} A systematic, uncertainty-aware framework that quantifies the economic value of adopting AI solutions in healthcare by combining FL--TDABC costing with fuzzy logic and ROI analysis.
    \item \textbf{Lifecycle costing of AI adoption.} An extension of FL--TDABC beyond service delivery to the AI solution itself, classifying costs into one-time development or acquisition costs, recurring costs, and other project-specific costs.
    \item \textbf{Uncertainty-aware economic metrics.} Annual and cumulative ROI, net savings, and break-even measures that propagate fuzzy time and cost estimates, so that expected, optimistic, and pessimistic scenarios follow from the same model.
    \item \textbf{An early health technology assessment case study.} A demonstration of FLARE on AI-assisted large vessel occlusion detection in the CT stroke pathway for acute ischemic stroke, including a sensitivity analysis of how annual patient volume affects ROI and time to break-even.
\end{itemize}

\section{Background and Related Work}

\subsection{Prior Research on the Economic Evaluation of AI in Healthcare}

The need for economic evidence has been echoed across both academic studies and industry analyses. Bharadwaj et al. \cite{Bharadwaj2024} notes that “AI research has focused primarily on the accuracy of algorithms, with little evidence measuring the impact of these innovations on patient- and physician-relevant outcomes or the broader benefits to the settings within which they are deployed.”  
In other words, while AI models are often benchmarked for predictive performance, there is limited evidence quantifying how these models translate into time savings, workflow improvements, or measurable cost reductions.  
The authors argue that there is an urgent need for Return on Investment (ROI) calculators that can capture and communicate the real-world benefits that radiology AI solutions bring to multiple stakeholders, clinicians, administrators, and patients alike.

Similarly, Rayscape \cite{Rayscape2024}, a radiology AI company providing workflow optimization and decision-support tools, emphasizes that many commercial ROI claims are incomplete because they only account for the algorithm’s purchase or deployment cost.  
Their report warns that “many AI vendors offering single-point solutions fail to accurately represent the true costs involved, often providing an ROI calculation based solely on the cost of deploying the algorithm itself, neglecting the broader investment made by the deploying organization.”  
These hidden costs include integration into existing systems, IT infrastructure, data preparation, clinician evaluation, and continuous performance monitoring.  
Rayscape, therefore, calls for ROI assessments that reflect the full scope of adoption, treating AI as an organizational transformation rather than a plug-in product. 

Other studies reinforce this concern, noting that economic analyses of AI adoption remain scarce despite rapid technical progress. For instance, Fasterholdt et al. \cite{fasterholdt2022value} highlight that most medical imaging studies focus on retrospective performance evaluation rather than assessing clinical or economic value. In contrast, Kastrup et al. \cite{kastrup2024landscape} describe the limited evidence for AI’s cost-effectiveness in healthcare more broadly. Most recently, Zhang et al. \cite{zhang2025economic} emphasize that while AI promises diagnostic and operational benefits in healthcare—including neurological applications—its deployment introduces substantial financial commitments related to data infrastructure, personnel training, and ongoing system maintenance.

\subsection{Why TDABC Emerges?}

To evaluate the economic value of AI adoption, it is necessary to understand how costs are computed in healthcare. Bottom-up costing approaches are particularly useful when the goal is to estimate costs at the activity or patient level by tracing the resources consumed, including staff time, equipment use, and consumables \cite{leusder2022cost}\cite{martin2018using}\cite{ninerola2021improving}. Among these, time-driven activity-based costing (TDABC) has been identified as especially valuable in healthcare because it enables detailed process mapping, resource-level costing, and greater transparency in estimating the cost of care delivery, particularly in value-based healthcare contexts \cite{keel2017time}\cite{kaplan2011solve}.

Traditional costing systems, often based on volume or departmental allocations, were criticized for their inability to reflect the true resource consumption of complex services such as healthcare \cite{drucker2012management}\cite{FuzzyTDABC2013}. 
They typically allocate overheads in proportion to direct labor or material costs, masking inefficiencies and failing to account for non-value-adding activities \cite{carroll2016growing}. 
To address these limitations, ABC was developed in the 1980s by Kaplan and Bruns \cite{kaplan1987accounting}, building on Staubus’ input–output accounting principles \cite{staubus1971activity}. 
ABC sought to trace costs more accurately by linking resource use to activities rather than to broad cost centers \cite{popesko2013specifics}. 
However, ABC has been criticized for being data- and labor-intensive to implement, costly to maintain, and vulnerable to subjectivity because activity definitions and cost-driver estimates often rely on staff interviews and survey-based judgments \cite{kaplan2007time}.

Kaplan and Anderson subsequently introduced the \textit{TDABC} method as a simplified and scalable evolution of ABC \cite{kaplan2007time}. 
Rather than relying on dozens of activity drivers, TDABC replaces them with two core parameters: 
(i) the \emph{cost of supplying capacity, i.e, Capacity Cost Rate (CCR)} and (ii) the \emph{time required to perform each activity}.  
This made it possible to compute the cost of a service using the simple formula:

\begin{equation}
\text{Activity Cost} = \text{Time to Perform Activity} \times \text{Capacity Cost Rate (CCR)},
\end{equation}

The CCR is obtained by dividing the total cost of a resource group by its practical capacity. The practical capacity is the portion of total available time remaining after deducting non-productive hours, such as meetings, sick leave, and breaks.

The move from ABC to TDABC addressed several long-standing issues:
\begin{itemize}
    \item \textbf{Reduced complexity and maintenance effort:} Unlike ABC, TDABC does not require repeated interviews or recalibration of cost drivers each time workflow conditions change.
    \item \textbf{Improved treatment of idle capacity:} TDABC explicitly incorporates \emph{practical capacity}, allowing for more realistic cost attribution.
    \item \textbf{Dynamic adaptability:} Because time equations can be updated easily, TDABC adapts better to evolving processes, new technologies, or AI-assisted workflows, making it ideal for modern healthcare operations.
\end{itemize}

\begin{figure}[htbp]
    \centering
    \includegraphics[width=0.9\textwidth]{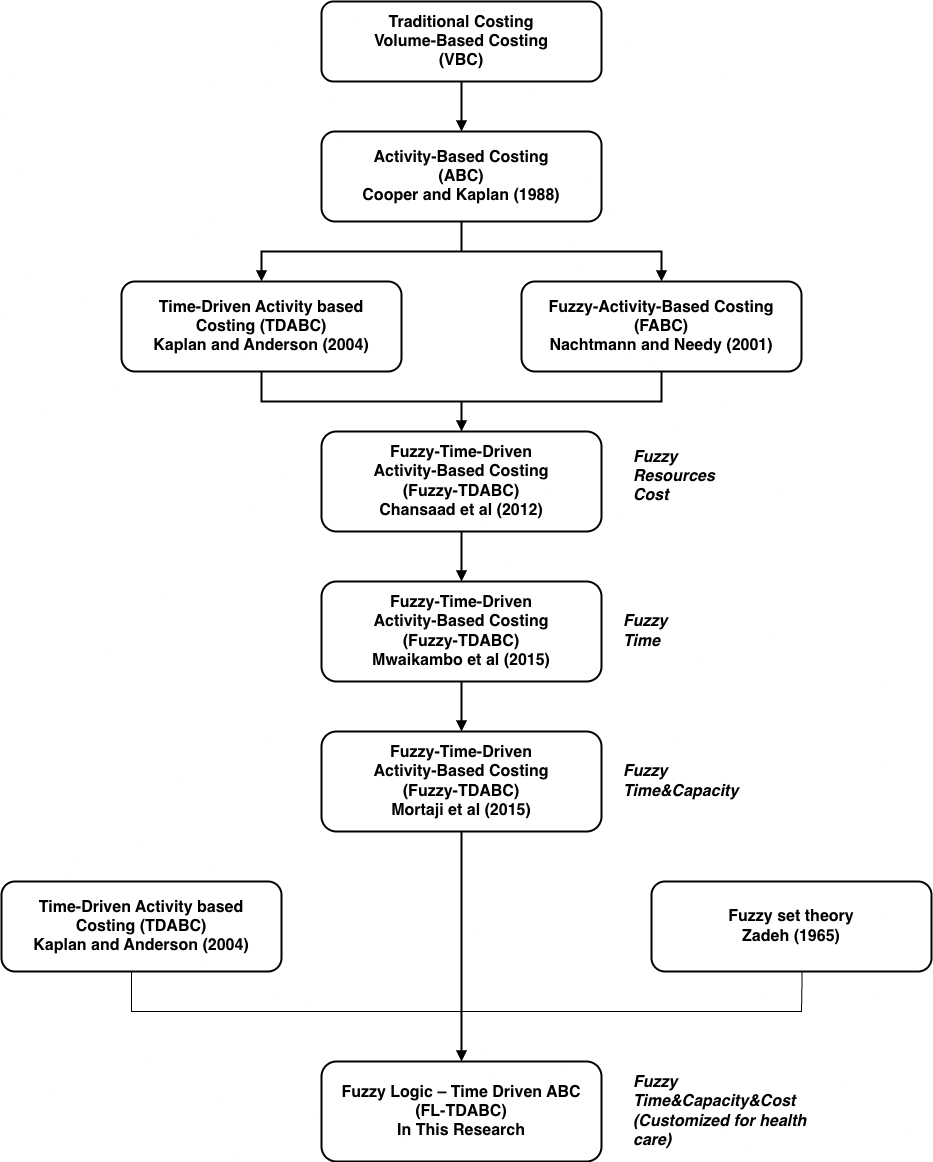}
    \caption{Evolution of costing methodologies from traditional systems to fuzzy logic–enhanced approaches.
    Adapted from Ostadi et al. \cite{Ostadi2019}}
    \Description{Diagram tracing the progression of costing methods in four stages: traditional volume- and department-based costing, activity-based costing (ABC), time-driven activity-based costing (TDABC), and fuzzy logic-enhanced FL-TDABC. Each stage is annotated with the limitation of the preceding method that motivated it.}
    \label{fig:evolution_costing_methods}
\end{figure}

These advantages explain why TDABC has become the preferred costing methodology across empirical healthcare studies. 
Its capacity to quantify both \emph{time} and \emph{cost} per activity provides a transparent, reproducible basis for evaluating efficiency improvements brought about by AI adoption. The progression from traditional costing systems to fuzzy logic–enhanced approaches is summarized in Figure~\ref{fig:evolution_costing_methods}.

\subsection{Why Fuzzy Logic Complements TDABC?}

While TDABC lays the foundation for estimating costs based on time and capacity, it assumes that the time required to complete each activity can be represented by a single fixed value.  
In real-world healthcare settings, however, activity durations and cost of resources vary widely due to patient complexity, workflow interruptions, and human judgment.  
Consequently, relying on fixed point estimates introduces bias and uncertainty into cost calculations, particularly when those estimates are derived from limited observations or expert opinion.

Koster et al. \cite{koster2023dealing} identify this as a key limitation in both ABC and TDABC.  
In their analysis of the Dutch Diagnosis Treatment Combination (DTC) reimbursement system, time estimates used to determine hospital payments are undisclosed and treated as “black boxes.”  
Because these point estimates mask the underlying variation in time spent by medical staff across patients and conditions, they hinder cost transparency.  
Even TDABC, though more granular than ABC, remains vulnerable to the same bias, since it relies on average time inputs that do not capture the full distribution of possible values.

To address this limitation, recent studies integrate fuzzy logic (FL) into TDABC, forming the hybrid \textit{Fuzzy Logic Time-Driven Activity-Based Costing (FL-TDABC)} model \cite{koster2023dealing}\cite{FuzzyTDABC2013}\cite{Ostadi2019}.
Fuzzy logic, introduced by Zadeh in 1965, provides a mathematical framework for representing imprecise or uncertain information \cite{koster2023dealing}\cite{zadeh1996fuzzy}. Imprecise information in healthcare is predominantly the result of variability in the time spent on procedures \cite{koster2023dealing}.
Unlike classical logic, which categorizes values as entirely true or false, fuzzy logic allows intermediate degrees of membership between 0 and 1.  
This makes it well-suited to healthcare, where time and resource utilization cannot be known with complete certainty but can be described using linguistic or approximate values such as “minimum,” “most likely,” and “maximum.”

\subsection{Fuzzy Numbers and Their Role in Cost Estimation}
The concept of fuzziness originates from Zadeh’s foundational work on fuzzy set theory \cite{zadeh1996fuzzy}, which models imprecise information using degrees of membership between 0 and 1 instead of binary true/false values.  
In the context of cost accounting, this means that parameters such as time can be treated as uncertain but bounded quantities rather than fixed constants.

In FL-TDABC, uncertainty in activity time estimates is represented through \textit{Triangular Fuzzy Numbers} (TFNs), each defined by three parameters:  
$(a, b, c)$, where $a$ represents the minimum possible time, $b$ the most likely time, and $c$ the maximum time.  
The defuzzified or expected time, $x^*$, is calculated as:

\begin{equation}
x^* = \frac{a + b + c}{3}.
\end{equation}

These TFNs form a spectrum of plausible durations rather than a single deterministic value.  
By incorporating fuzzy numbers, FL-TDABC captures both optimistic and pessimistic scenarios and generates a more realistic expected cost outcome.  
As shown by Mortaji et al. \cite{FuzzyTDABC2013}, this approach allows decision-makers to quantify uncertainty explicitly, reducing the risk of over- or underestimating costs when time data are incomplete or subjective.

Fuzzy modeling also enables sensitivity analysis without re-running the entire cost model: lower-bound (best-case) and upper-bound (worst-case) estimates can be obtained directly by substituting $a$ or $c$ in place of $x^*$.  
This provides decision-makers with a bounded view of potential cost variation, improving confidence in the robustness of ROI projections.

\subsection{Why Fuzziness Matters for Decision-Making?}

In practice, fuzziness represents more than mathematical uncertainty; it reflects the reality of decision-making in complex systems like healthcare.  
Time spent per activity fluctuates with patient mix, team composition, and workflow design; resource costs shift with policy changes and technology upgrades.  
By modeling these variations rather than ignoring them, FL-TDABC and FLARE align cost and ROI estimation with real-world variability.  
This makes the framework more credible to practitioners, policymakers, and investors who must make high-stakes decisions under uncertainty.

Therefore, fuzzy logic complements TDABC by transforming a deterministic costing system into a probabilistic, context-aware decision-support tool.  
This integration enhances cost transparency and improves the robustness of cost estimation by explicitly accounting for variation in activity durations and resource use, while also providing a stronger foundation for ROI computation within the FLARE framework.

\subsection{Understanding Return on Investment }

It is essential to understand the concept of ROI and its relevance in healthcare decision-making. ROI is a financial metric used to evaluate the efficiency or profitability of an investment by comparing the benefits gained with the costs incurred \cite{botchkarev2011return}. It is generally expressed as:

\begin{equation}
ROI = \frac{\text{Total Benefits} - \text{Total Costs}}{\text{Total Costs}} \times 100\%.
\end{equation}

In healthcare, ROI helps determine whether a particular intervention, technology, or program provides sufficient value to justify its cost \cite{masters2017return}\cite{thusini2022development}. While ROI is widely used in business and industry, its application in healthcare has grown rapidly as health systems seek evidence of both clinical and financial impact before adopting innovations \cite{turner2023analyses}. 

An ROI analysis in healthcare typically considers both \textit{direct} and \textit{indirect} elements \cite{turner2023analyses}. Direct components include measurable financial outlays such as equipment, staff time, infrastructure, and software costs. Indirect components capture broader benefits like reduced treatment times, improved diagnostic accuracy, or fewer readmissions. These indirect gains, though harder to quantify, often represent the real value that innovative technologies such as AI bring to healthcare systems.

However, traditional ROI assessments often face limitations. Many use fixed, deterministic inputs and fail to account for uncertainty or variation in patient volume, staff efficiency, or workflow design \cite{turner2023analyses}. Others focus narrowly on acquisition costs while ignoring integration, maintenance, and long-term lifecycle expenses \cite{Bharadwaj2024}\cite{Rayscape2024}.

As hospitals and policymakers move toward value-based healthcare, there is growing recognition that ROI must evolve beyond static financial ratios to reflect operational dynamics and uncertainty \cite{thusini2022development}. This is particularly important in evaluating AI solutions, where costs and benefits unfold over time and depend on complex inter-dependencies among staff, technology, and patient outcomes \cite{khanna2022economics}. Hence, evaluating the ROI of AI solutions demands a holistic framework that captures both the variability of activities involved in delivering a service and the full range of costs incurred across the lifecycle of AI adoption.

Alternative economic evaluation frameworks exist, including Net Present Value (NPV) \cite{shou2022literature}, Internal Rate of Return (IRR) \cite{mellichamp2017internal}, and cost-effectiveness or cost–utility analysis \cite{robinson1993cost}. 
While these approaches are valuable in formal health economic evaluations, they require long-term outcome data, explicit discounting assumptions, and utility estimates that may not be reliably available at early stages of technology development \cite{girling2010early}. In practice, simpler financial metrics such as ROI can provide a more accessible and transparent first step in evaluating early investments, particularly when detailed health outcome data are lacking \cite{turner2023analyses}. As such, ROI was selected as the primary economic metric in FLARE as an initial decision-support indicator, with the understanding that comprehensive economic evaluation (e.g., cost–utility analysis) remains important as more data become available \cite{ijzerman2017emerging}.

\subsection{Research Gap}

Taken together, the studies reviewed above reveal a key gap: existing approaches fail to provide a comprehensive, transparent accounting of the costs and benefits associated with adopting AI solutions throughout their full lifecycle.
This lifecycle perspective is crucial because, as highlighted in the literature, most AI evaluations stop at model accuracy or deployment cost.
Addressing this gap requires a framework capable of tracing resource use and time commitments not only during service delivery, but also during the development, validation, deployment, and operational phases of an AI system.

\subsection{Design Requirements for ROI (Granularity, Data Provenance, and Reporting Core)}

Having established TDABC as the preferred costing method, the next step is to define how it supports the computation of ROI for adopting technological/AI solutions in healthcare.  
While TDABC provides the foundation for estimating the true cost of delivering clinical services, the goal of this research extends further:  
To estimate the ROI of adopting an AI solution across its entire lifecycle, from development and validation to deployment, integration, and sustained use within healthcare workflows.
TDABC, with its activity-level structure and time-based logic, provides the ideal foundation for this.

Building on this foundation, the design requirements for an ROI framework can be summarized across three interdependent domains:

(1) \textbf{Granularity of cost estimation}, which requires the framework to operate at the activity level, tracing time and resource use across both the clinical workflow and the AI adoption pathway, including development, validation, training, integration, and maintenance costs, each captured through TDABC’s time- and capacity-based logic; 

(2) \textbf{Data provenance and uncertainty handling}, which acknowledges that healthcare operations and AI development pipelines are inherently uncertain and therefore require explicit treatment of variability in activity times, staff utilization, and practical capacity; and 

(3) \textbf{Reporting and integration}, which demands transparent, stakeholder‑oriented outputs such as time‑indexed ROI, multi‑year horizons, break‑even points, and sensitivity analyses to help decision‑makers understand when and how AI investments become economically sustainable.

\section{Methods}

\subsection{FLARE: Service Costing}

FL-TDABC retains the seven foundational steps proposed by Kaplan and Porter \cite{kaplan2011solve} but replaces deterministic time inputs with fuzzy estimates, allowing uncertainty to be modeled explicitly.

\begin{figure}[htbp]
    \centering
    \includegraphics[width=0.85\textwidth]{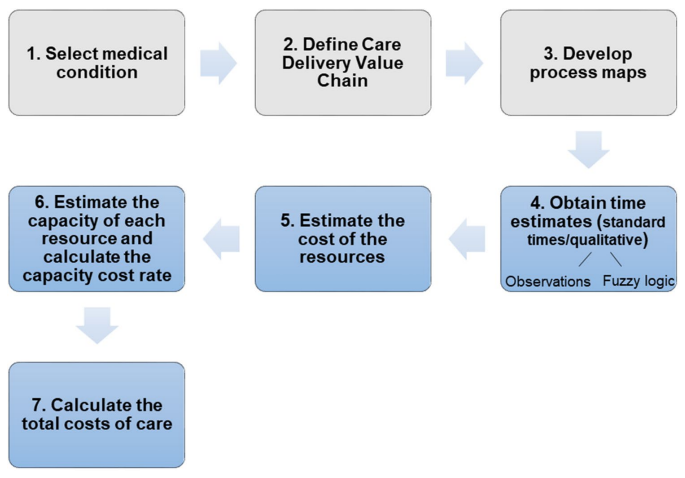}
    \caption{Seven-step FL-TDABC framework integrating fuzzy time estimates within the TDABC cost model. Obtained from Koster et al.\cite{koster2023dealing}}
    \Description{Flow diagram of the seven sequential FL-TDABC steps: select the medical condition or service, define the care delivery value chain, develop process maps, obtain time estimates as triangular fuzzy numbers, estimate resource costs, compute the capacity cost rate, and calculate the total fuzzy cost of care.}
    \label{fig:fl_tdabc_steps}
\end{figure}

\paragraph{Step 1: Select the Medical Condition or Service}
Define the clinical or operational process to be costed (e.g., ischemic stroke care, MRI workflow).  
This sets the boundary for the costing model and ensures that subsequent steps capture all relevant activities.

\paragraph{Step 2: Define the Care Delivery Value Chain}
Map the end-to-end stages of the service, identifying where value is created or transferred between departments or resources.  
This stage establishes the high-level structure for subsequent activity mapping.

\paragraph{Step 3: Develop Process Maps}
List all activities required within each stage and the resources performing them (e.g., clinicians, technicians, machines).  
In healthcare, this step often draws on process observation, protocol review, or time–motion studies.

\paragraph{Step 4: Obtain Time Estimates under Uncertainty}
Unlike traditional TDABC, which assigns a fixed duration to each activity, FL-TDABC represents each time estimate as a \textbf{Triangular Fuzzy Number (TFN)} $(a, b, c)$, where:
\[
a = \text{minimum time}, \quad
b = \text{most likely time}, \quad
c = \text{maximum time}.
\]
The defuzzified or expected time is computed as:
\[
x^{*} = \frac{a + b + c}{3}.
\]
This approach captures the variability that arises from patient complexity, workflow interruptions, and differences in staff efficiency.  
It allows decision-makers to propagate uncertainty throughout the costing model rather than relying on a single deterministic value.

\paragraph{Step 5: Estimate Resource Costs}
For each resource, determine the full cost, including salaries, equipment depreciation, consumables, utilities, and overheads, over a fixed period (e.g., annually).

\paragraph{Step 6: Estimate Capacity and Compute the Capacity Cost Rate (CCR)}
The \textit{Capacity Cost Rate} quantifies the cost per unit of productive time for each resource:
\begin{equation}
CCR_i = \frac{\text{Annual Cost of Resource } i}{\text{Practical Capacity of Resource } i}.
\end{equation}
Practical capacity is typically defined as total available time minus non-productive time (e.g., breaks, meetings, leave).

\paragraph{Step 7: Calculate Total Fuzzy Cost of Care}
Finally, the total cost per service is computed by multiplying each activity’s fuzzy time estimate by the corresponding resource CCR:
\begin{equation}
Cost_{service} = \sum_{i=1}^{n} CCR_i \times x_i^{*}.
\end{equation}
This produces not only an expected cost but also lower and upper bounds:
\begin{equation}
Cost_{low} = \sum CCR_i \times a_i, \quad
Cost_{high} = \sum CCR_i \times c_i,
\end{equation}
Corresponding to optimistic and pessimistic scenarios. 

The result represents the cost of completing a \textbf{single service cycle}—that is, one full iteration of the process being costed from start to finish.  
In healthcare, this would correspond to the cost of care for one patient, one diagnostic procedure, or one treatment episode.  
In manufacturing, it could represent the cost of producing a single unit, while in administrative or digital services, it may correspond to one transaction or process instance.  ,
Once the cost of a single service cycle is known, it can be scaled to estimate total or annual costs by multiplying by the number of cycles completed in a given time period:
\begin{itemize}
    \item In healthcare: cost per patient $\times$ annual patient volume = annual care cost.
    \item In manufacturing: cost per unit $\times$ annual production volume = annual production cost.
    \item In service industries: cost per transaction $\times$ annual transactions = total annual operating cost.
\end{itemize}

\noindent
This integration transforms TDABC from a static cost-allocation tool into a dynamic, uncertainty-aware decision-support framework.  
By combining the transparency of TDABC with the flexibility of fuzzy logic, FL-TDABC enables more robust cost estimation in healthcare contexts where time data are subjective, variable, or incomplete.  
Within the FLARE framework, this serves as the computational backbone for modeling the ROI of AI-integrated workflows, ensuring that both deterministic and uncertain elements of cost and time are explicitly represented.

\subsection{FLARE: AI Solution Costing}
Building on the cost of service delivery, the next component of FLARE quantifies the total cost of the AI solution across its life-cycle. These vary depending on whether an organization is developing 
its own model, adopting an external solution, or paying for a SaaS (Software as a Service) platform. 
To capture these differences, we classify costs into three categories: 
(i) one-time development or acquisition costs, 
(ii) recurring costs, and 
(iii) other project-specific costs.

\subsubsection{One-time Development Costs}

This subsection covers in-house development, where an organization develops its own model.

From the cost of service using FL-TDABC, we can determine the cost to treat a single patient, and by scaling with annual volume we obtain the annual cost of care.  
Using the same logic for development, we ask: \textit{“How much does one sample of data cost?”}  
Once we know the per-sample cost, we can scale to the dataset size to estimate the total development cost.

Applying FL-TDABC to AI development describes model building as a series of measurable, time-driven activities.  
Each step consumes resources (human and non-human) that we can estimate using the same principles.  
Activities depend on the specific machine learning problem (supervised, semi-supervised, self-supervised) and the chosen workflow.

\paragraph{Step 1: Define the Service (AI Development Unit)}

The unit of analysis is a single data sample processed during development.  
A sample passes through activities such as data preparation, labeling (if applicable), training, and evaluation.  
Per-sample cost can be multiplied by \(N\) (or by cohort-specific sizes) to obtain total development cost.

\paragraph{Step 2: Define the Value Chain}

Activities depend on the problem being solved and the data used.  
For supervised learning, there are usually distinct training/validation and test phases; semi/self-supervised learning may rely on unlabeled data.  
As an \emph{illustrative example} (not a prescription), a supervised workflow may include:
\begin{enumerate}
    \item \textbf{Data Gathering} (\(A_{dg}\)) – collecting scans/datasets/inputs.
    \item \textbf{Preprocessing \& Pipeline Setup} (\(A_{pp}\)) – cleaning, normalization, formatting.
    \item \textbf{Labeling \& Annotation} (\(A_{l}\)) – expert or technician labeling (if required).
    \item \textbf{Model Training} (\(A_{t}\)) – training runs, tuning, and job management.
    \item \textbf{Evaluation \& Verification} (\(A_{e}\)) – model evaluation on validation or test sets, and/or human verification.
\end{enumerate}
In practice, you should chart only the activities that occur for the specific project.

\paragraph{Step 3: Develop Process Maps}

Link each activity to the resources it uses:
\begin{itemize}
    \item \textbf{Human:} data/ML engineers, annotators, clinicians, researchers.
    \item \textbf{Computational:} servers, GPUs/CPUs, cloud instances, storage.
    \item \textbf{Software/Materials:} licenses, APIs, specialized toolkits.
\end{itemize}
This keeps costing transparent and modular.

\paragraph{Step 4: Obtain Fuzzy Time Estimates}

Represent per-sample activity times with triangular fuzzy numbers:
\[
(a,b,c)_{dg},\; (a,b,c)_{pp},\; (a,b,c)_{l},\; (a,b,c)_{t},\; (a,b,c)_{e},
\]
where \(a\) = minimum, \(b\) = most likely, \(c\) = maximum (minutes), and expected time
\[
x^{*} = \frac{a+b+c}{3}.
\]

\paragraph{Step 5: Estimate Resource Rates}

Assign an hourly or usage-based rate to each resource (e.g., \$/hour for staff, \$/GPU-hour, \$/GB-hour).  
Include overheads if appropriate.

\paragraph{Step 6: Compute Capacity Cost Rate (CCR)}

Convert each rate to a per-minute cost:
\begin{equation}
CCR_i = \frac{\text{Hourly Rate}_i}{60}.
\end{equation}

\paragraph{Step 7: Calculate Per-Sample Development Cost \;(illustrative)}

For this \emph{example} activity set, the per-sample expected cost is:
\begin{equation}
\begin{aligned}
C_{dg}^{*} &= \sum_i CCR_i \cdot x^{*}_{dg,i}, \\
C_{pp}^{*} &= \sum_i CCR_i \cdot x^{*}_{pp,i}, \\
C_{l}^{*}  &= \sum_i CCR_i \cdot x^{*}_{l,i}, \\
C_{t}^{*}  &= \sum_i CCR_i \cdot x^{*}_{t,i}, \\
C_{e}^{*}  &= \sum_i CCR_i \cdot x^{*}_{e,i}.
\end{aligned}
\end{equation}
\begin{equation}
C_{\text{sample}}^{*} = C_{dg}^{*} + C_{pp}^{*} + C_{l}^{*} + C_{t}^{*} + C_{e}^{*} + C_{\text{other}},
\end{equation}
Where \(C_{\text{other}}\) captures additional costs not tied to activity time (e.g., license fees, data purchase, admin, infrastructure maintenance).

\paragraph{Scaling to Dataset Level (cohort-specific or pooled)}

If all samples undergo the same activities, the total in-house development cost is:

\begin{equation}
AI\_DevCost^{*} = N \cdot C_{\text{sample}}^{*}.
\end{equation}
When activities differ by cohort (e.g., training/validation/test or unlabeled pools), scale each cohort separately and add:
\begin{align}
AI\_DevCost^{*}
&= N_{\text{train}} \cdot C_{\text{sample,train}}^{*}
 + N_{\text{val}} \cdot C_{\text{sample,val}}^{*} \nonumber \\
&\quad + N_{\text{test}} \cdot C_{\text{sample,test}}^{*}
 + N_{\text{unlab}} \cdot C_{\text{sample,unlab}}^{*} \nonumber \\
&\quad + C_{\text{other}}.
\label{eq:aidevcost_cohort}
\end{align}
Fuzzy bounds follow by substituting \(a\) and \(c\) for \(x^{*}\) in each term.

\noindent
In summary, FL-TDABC for in-house development answers \textit{“How much does one sample cost to process?”}.  
By charting the true activities and resources for the specific problem, and then scaling by the relevant cohort sizes, we obtain a transparent and adaptable estimate of development cost under uncertainty.

\subsubsection{Recurring and Other Costs}

These are the ongoing costs of operating and maintaining the AI solution. 
Rather than treating them as a single annual value, they can be expressed at a finer
time resolution (e.g., monthly), since salaries, infrastructure, or SaaS fees may change during the year. 
\begin{itemize}
    \item $C_{infra,t,m}$: infrastructure costs in year $t$, month $m$ (e.g., servers, GPUs, storage, networking).
    \item $C_{mgmt,t,m}$: management and maintenance costs in year $t$, month $m$ (e.g., DevOps, retraining, monitoring).
    \item $C_{SaaS,t,m}$: subscription or licensing fees in year $t$, month $m$.
    \item $C_{other,t,m}$: other recurring costs in year $t$, month $m$ not captured above.
\end{itemize}

For each year $t$ and month $m \in \{1,\dots,12\}$, we define:
\begin{equation}
    AI\_Cost_{t,m} = C_{infra,t,m} + C_{mgmt,t,m} + C_{SaaS,t,m} + C_{other,t,m}
\end{equation}

The annual recurring cost in year $t$ is then:
\begin{equation}
    AI\_{Recurring}^{(t)} = \sum_{m=1}^{12} AI\_Cost_{t,m}
\end{equation}

\paragraph{Other one-time project costs.}
These are additional costs not tied directly to development or monthly operations, 
such as software licenses or acquisition fees. If such costs occur in year $t$, they
are denoted as $C_{other}^{(t)}$.

\paragraph{Performance Reflection through Time.}
If the AI system performs poorly, the time required for verification increases. 
This verification time, priced through the resource’s CCR, translates naturally into additional cost:
\begin{equation}
Cost_{\text{verification}} = CCR_{\text{verifier}} \times Time_{\text{review}}.
\end{equation}
Thus, performance penalties are endogenously captured through time, maintaining consistency with the TDABC logic.

\subsection{FLARE: Benefit and ROI Metrics}

It is important to note that costs are not uniform across years.  
In Year~1, the ROI reflects both one-time development or acquisition costs and recurring operational expenses such as infrastructure, management, and SaaS fees.  
In Year~2 and beyond, if the model has already been developed, total costs generally decrease, consisting mainly of recurring expenditures unless additional one-time payments (e.g., major upgrades) are incurred.  
This means ROI in later years often appears higher than in Year~1, since benefits remain constant, or may even increase as usage scales, while costs decline. 

The total AI cost in year $t$ can then be defined as:
\begin{equation}
    AI\_Cost_{year\ t} =
    \mathbf{1}\{t=1\}\cdot AI\_DevCost^{*}
    + AI\_{Recurring}^{(t)}
    + C_{other}^{(t)},
\end{equation}
where the indicator function $\mathbf{1}\{t=1\}$ is defined as:
\[
    \mathbf{1}\{t=1\} =
    \begin{cases}
        1, & \text{if } t = 1, \\
        0, & \text{if } t \neq 1.
    \end{cases}
\] This ensures that one-time development costs are only included in the first year.

The benefit in month $m$ of year $t$ is defined as the difference in unit cost per service cycle (without vs.\ with AI) multiplied by the monthly volume:
\begin{equation}
    Benefit_{t,m} = \big(UC^{noAI}_{t,m} - UC^{AI}_{t,m}\big)\, V_{t,m},
\end{equation}

\begin{itemize}
    \item $UC^{noAI}_{t,m}$ – unit cost per service cycle without AI in month $m$ of year $t$,
    \item $UC^{AI}_{t,m}$ – unit cost per service cycle with AI in month $m$ of year $t$,
    \item $V_{t,m}$ – number of service cycles (volume) in month $m$ of year $t$.
\end{itemize}

And the annual benefit is:
\begin{equation}
    Benefit_{year\ t} = \sum_{m=1}^{12} Benefit_{t,m}.
\end{equation}
Additionally, this monthly expansion allows FLARE to capture intra-year variability such as salary increases in Month~3, subscription adjustments in Month~6, or seasonal fluctuations in service volume, while still reporting ROI at the annual or multi-year level.

ROI can be expressed annually, which answers the question: “How much did the AI save me this year?”:
\begin{equation}
    ROI_{year\ t} = \frac{Benefit_{year\ t} - AI\_Cost_{year\ t}}{AI\_Cost_{year\ t}}
\end{equation}

Alternatively, ROI can be defined as a cumulative measure that grows over time, capturing all historical costs and benefits since deployment. This is vital when decision-makers want to ask: “How much has the AI saved me since adoption?”
\begin{equation}
    ROI_{cumulative,\ T} =
    \frac{\sum_{t=1}^{T} Benefit_{year\ t} - \sum_{t=1}^{T} AI\_Cost_{year\ t}}
         {\sum_{t=1}^{T} AI\_Cost_{year\ t}}
\end{equation}

This distinction allows decision-makers to evaluate short-term efficiency and long-term financial sustainability within the same analytical framework.

\subsection{Early Health Technology Assessment (EHTA) Case Study}

We demonstrate the practical application of the proposed FLARE framework through a real-world EHTA scenario. In this case study, a hospital is evaluating whether to adopt an AI solution for large vessel occlusion (LVO) detection within its CT imaging pathway for acute ischemic stroke (AIS). The goal is to understand whether the investment would be economically viable over time, given realistic uncertainty in costs, workload, and operational efficiency.

The analysis focuses on two objectives:
\begin{enumerate}
    \item estimating the cost of delivering care along the conventional CT pathway, and
    \item assessing the economic impact and ROI of integrating an AI solution that detects LVO from CTA.
\end{enumerate}

The purpose of this case study is not to produce exact cost figures. Due to the limited availability of public institutional and financial data, precise cost estimation is not feasible. Instead, the aim is to \textbf{demonstrate how the FLARE framework systematically structures and quantifies each component of AI adoption and healthcare delivery in a transparent, reproducible way to support informed decision-making}. By explicitly mapping clinical activities, assigning time and resource estimates, and incorporating uncertainty through fuzzy logic, the framework illustrates how early-stage economic evaluations can be performed. By modeling both the existing service and the AI-enhanced workflow, we show how FLARE can support early decision-making on AI adoption, helping stakeholders (clinicians, hospital executives, and policymakers) understand potential value, resource implications, and ROI.

Because EHTA is intended to inform adoption decisions, it becomes meaningful only once a solution has demonstrated sufficient technical validity to plausibly influence clinical decision-making. If an algorithm does not meet minimum performance expectations for clinical use, there is little justification for pursuing downstream economic evaluation, as the technology would not be considered a viable candidate for adoption regardless of cost or efficiency gains. Accordingly, this case study is based on an AI solution for LVO detection that has already been established and validated in the existing literature \cite{brugnara2023deep}. This allows the analysis to focus on a realistic adoption scenario in which a high-performing AI tool is available, and the central question becomes whether translating such technical performance into clinical practice is economically viable.

To ensure that the modeled workflow and time estimates were clinically realistic, expert input was incorporated during the case study design. 
A radiologist at Foothills Medical Centre (Calgary) served as the primary clinical collaborator during the time–motion study, providing activity definitions and fuzzy time estimates for the CT stroke pathway.

This input was used to ground the case study in real clinical practice rather than to formally validate the economic outputs. 
Nevertheless, the costing logic underlying FLARE is based on TDABC and fuzzy logic, both of which are well established and widely validated in prior academic and industry work \cite{koster2023dealing} \cite{keel2017time}.

LVO stroke is a type of ischemic stroke caused by blockage of a major intracranial artery and is associated with severe neurological deficits and high disability if not treated promptly \cite{rennert2019epidemiology}. Rapid imaging-based diagnosis using NCCT and CTA is essential to identify eligible patients for EVT, the standard of care for LVO cases \cite{abdalkader2023neuroimaging}. In many hospitals, patients follow the conventional CT pathway—presenting through the emergency department (ED), receiving NCCT and CTA in the CT suite, and then being transferred to the neuro-interventional radiology (NIR) suite for EVT if indicated \cite{sangha2024time}. An alternative workflow, the direct-to-angiography pathway, bypasses the CT suite to shorten imaging-to-reperfusion time but can lead to higher costs when patients are not EVT-eligible. A recent TDABC study \cite{sangha2024time} compared these pathways, showing that although the direct approach reduces treatment delay, the CT pathway remains more cost-efficient when full eligibility screening is required.

\begin{figure}[t]
  \centering
  \includegraphics[width=\textwidth,height=0.85\textheight,keepaspectratio]{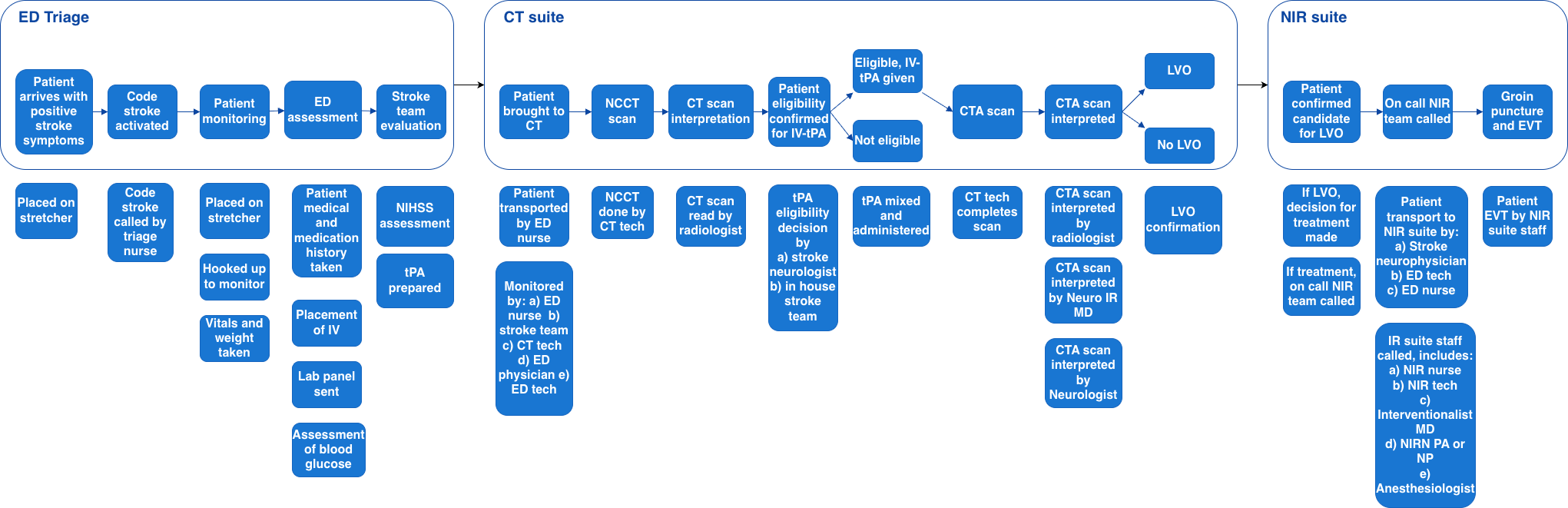}
  \caption{CT pathway for acute ischemic stroke patients with suspected LVO, obtained from Sangha et al.}
  \Description{Flowchart of the conventional CT stroke pathway across three stages. In ED triage the patient is placed on a stretcher, a code stroke is called, and the stroke team performs an NIHSS assessment. In the CT suite an NCCT and then a CTA are acquired and interpreted, and IV-tPA eligibility and the presence of a large vessel occlusion are confirmed. Eligible patients are transferred to the neuro-interventional radiology suite for groin puncture and endovascular thrombectomy.}
  \label{fig:ct_pathway}
\end{figure}

Advances in medical imaging AI have made it possible to detect vascular occlusions automatically from CTA scans within minutes. Brugnara et al. \cite{brugnara2023deep} developed a deep learning system capable of identifying LVO and medium vessel occlusion (MeVO) with over 87\% sensitivity, achieving comparable or superior accuracy to current FDA-approved solutions. Integrating such AI tools into the CT pathway could speed up radiologist interpretation, reduce decision-making time, and optimize workflow efficiency by automatically flagging suspected LVO cases for priority review.

However, while performance metrics are well-documented, the economic value of implementing such AI systems in hospital workflows remains unclear. Quantifying this value—especially early in adoption—is essential for understanding whether time savings translate into meaningful financial and operational benefits.

\section{Results}

\subsection{Cost of the CT Pathway}
\label{sec:ct_cost_model}

This section estimates the cost of performing one complete CT stroke pathway (i.e, cost per patient) using FL-TDABC.  
The CT pathway begins at the emergency department (ED) triage and continues through CT/CTA imaging and interpretation, ending with possible endovascular thrombectomy (EVT) in the neuro-interventional radiology (NIR) suite for eligible patients.

The activity structure and resources were adapted from Sangha et al. \cite{sangha2024time}, who mapped the direct-to-angiography and CT-based workflows for ischemic stroke (Figure~\ref{fig:ct_pathway}).  
However, the study did not include time estimates for each activity.  
To fill this gap, we conducted a time study with a radiologist at Foothills Medical Centre (Calgary) to collect realistic fuzzy time estimates (minimum, most likely, and maximum) for each activity.  
Personnel hourly rates were obtained primarily from the Job Bank of Canada for Calgary, Alberta.  
When a specific role was unavailable, data were retrieved from ALIS Alberta or, as a last resort, from Glassdoor.  
This ensures that our cost estimates reflect current market rates in Calgary, Alberta, for healthcare.

The cost of each activity was then computed using the seven-step FL-TDABC framework. The full personnel rates, material costs, and detailed process map used for the FL--TDABC estimation of the CT pathway are provided in Appendix~\ref{appendix:ct_pathway_tables}.

A complete step-by-step methodological walkthrough of the FL--TDABC application for the CT pathway, including worked examples of fuzzy time calculations and activity costing, is presented in Appendix~\ref{appendix:ct_fltdabc_steps}.

Summing across all activities yields the per-patient expected, low, and high cost estimates:
\[
\text{Cost}_{\text{expected}} = 12099.45\text{ CAD}, \quad
\text{Cost}_{\text{low}} = 11491.47\text{ CAD}, \quad
\text{Cost}_{\text{high}} = 13011.22\text{ CAD}.
\]

According to Alberta Health Services (AHS), approximately 5{,}000 Albertans experience a stroke each year \cite{AHS_stroke}.  
Hence, scaling the per-patient cost to the provincial level gives an estimated annual CT pathway cost of:

\[
C_{\text{annual}}^{\text{expected}} = 60{,}497{,}250 \text{ CAD}, \quad
C_{\text{annual}}^{\text{low}} = 57{,}457{,}250 \text{ CAD}, \quad
C_{\text{annual}}^{\text{high}} = 65{,}056{,}100 \text{ CAD.}
\]

\subsection{Cost of AI Adoption for LVO Detection}
\label{sec:ai_cost_model}

This section applies FL--TDABC to estimate the cost of adopting an AI solution designed to automatically detect large vessel occlusion (LVO) on CTA scans.  
The information used here is primarily drawn from Brugnara et al. \cite{brugnara2023deep}, who developed and validated a deep learning model for LVO detection using CTA data from multiple clinical sites.  
Where data were not explicitly provided, informed assumptions were made based on literature and expert consultation, consistent with the purpose of this study—to demonstrate how FLARE can support early cost evaluation under uncertainty rather than to produce exact financial estimates. The detailed personnel rates and process maps for both the training and test cohorts are provided in Appendix~\ref{appendix:ai_costing}.

\subsubsection{Development Costs: ``How much does a sample of data cost?''}

Here we extend the same logic used in the CT pathway: “How much does one service cycle cost?”; analogously, for development we ask, \emph{“How much does one data sample cost to process through the development workflow?”}  
Not all samples undergo the same activities. To capture real resource use, we split samples by the actual activities they receive.  
Brugnara \textit{et al.}\ reported four cohorts; in this case study, we model the \emph{Heidelberg training cohort} and the \emph{Heidelberg internal test cohort}, because these have sufficient timing detail for the steps we cost. The complete seven-step FL--TDABC methodological walkthrough, including value-chain definition, activity mapping, fuzzy-time calculations, CCR estimation, and a complete worked example of per-activity costing, is provided in Appendix~\ref{appendix:ai_dev_fltdabc_steps}.

\begin{table}[h]
\caption{Cohorts used in Brugnara et al.}\label{tab:ai_cohorts}
\renewcommand{\arraystretch}{1.2}
\small
\begin{tabular*}{\textwidth}{@{\extracolsep\fill}p{0.34\textwidth}p{0.43\textwidth}p{0.17\textwidth}@{}}
\toprule
\textbf{Cohort} & \textbf{Description} & \textbf{Sample size ($N$)} \\
\midrule
Heidelberg training cohort & Internal data used for model training & 835 (71\%) \\
Heidelberg internal test cohort & Internal test data for validation & 344 (29\%) \\
External FAST cohort & Pseudo-prospective validation & Not reported \\
External UKB cohort & External pseudo-prospective validation & Not reported \\
\bottomrule
\end{tabular*}
\end{table}

\paragraph{Rates and CCR.}
All personnel hourly rates are for Calgary, Alberta (Obtained Job Bank of Canada / ALIS Alberta).  
Capacity cost rates (CCR) convert hourly rates to per-minute: $CCR = \text{Hourly}/60$.

\paragraph{Activity times (fuzzy).}
Where Brugnara \textit{et al.}\ report a range and a median, we use $(a,b,c)$ = (min, median, max).  
Where time is not reported, we use literature/expert estimates stated explicitly below:
\begin{itemize}
    \item \textbf{Data gathering}: Rava \textit{et al.}\ \cite{rava2021validation} report CTA acquisition of 3--5~min; we use $(a,b,c)=(3,4,5)$~min per sample as a proxy for data retrieval overhead.
    \item \textbf{Preprocessing}: Brugnara \textit{et al.}\ report $a=1.12$, $b=1.38$, $c=1.88$ minutes per sample.
    \item \textbf{Labeling/annotation}: Not reported; we use 22~min expected from Rava \textit{et al.} as a rough reference and set $(a,b,c)=(17,22,27)$~min (i.e., $\pm 5$ around 22).
    \item \textbf{Model training}: Not reported; we assume $(6,7,8)$~min \emph{per sample-equivalent} to price human time for job setup/monitoring by an ML engineer. Compute/GPU costs are excluded (as per the paper’s locally owned GPUs).
    \item \textbf{Inference}: Brugnara \textit{et al.}\ report $(a,b,c)=(0.27,0.33,0.47)$~min per sample. As we exclude GPU/compute pricing and assume no billed human oversight, we set its personnel cost to 0~CAD.
    \item \textbf{Evaluation/review}: Brugnara \textit{et al.}\ provide a range and median (values not numerically stated here); when numeric values are not available, we mark as \emph{NR} and exclude from the subtotal to avoid fabricating inputs.
\end{itemize}


\paragraph{Training cohort totals (scaled by $N{=}835$).}
\[
\begin{aligned}
C_{\text{train, low}} &= 835 \times 19.86 = 16{,}583.1\ \text{CAD},\\
C_{\text{train, expected}} &= 835 \times 25.17 = 21{,}058.7\ \text{CAD},\\
C_{\text{train, high}} &= 835 \times 30.59 = 25{,}584.4\ \text{CAD}.
\end{aligned}
\]

\paragraph{Test cohort totals (scaled by $N{=}344$).}
\[
\begin{aligned}
C_{\text{test, low}} &= 344 \times 19.32 = 6{,}646.08\ \text{CAD},\\
C_{\text{test, expected}} &= 344 \times 25.49 = 8{,}768.56\ \text{CAD},\\
C_{\text{test, high}} &= 344 \times 31.73 = 10{,}915.12\ \text{CAD}.
\end{aligned}
\]

The total one-time development cost of the AI model combines the contributions from the training and test cohorts.  
Each cohort’s cost is expressed at three levels—low, expected, and high—corresponding to the minimum, defuzzified, and maximum time scenarios derived from the triangular fuzzy estimates.

\[
AI\_DevCost =
C_{\text{train}} + C_{\text{test}},
\]

Where each component is evaluated at its respective fuzzy bound:

\[
\begin{aligned}
AI\_DevCost_{\text{low}} &= C_{\text{train,low}} + C_{\text{test,low}},\\
AI\_DevCost_{\text{expected}} &= C_{\text{train,expected}} + C_{\text{test,expected}},\\
AI\_DevCost_{\text{high}} &= C_{\text{train,high}} + C_{\text{test,high}}.
\end{aligned}
\]

Substituting the cohort totals:

\[
\begin{aligned}
AI\_DevCost_{\text{low}} &= 16{,}583.1 + 6{,}646.08 = 23{,}229.18\ \text{CAD},\\
AI\_DevCost_{\text{expected}} &= 21{,}058.7 + 8{,}768.56 = 29{,}827.26\ \text{CAD},\\
AI\_DevCost_{\text{high}} &= 25{,}584.4 + 10{,}915.12 = 36{,}499.52\ \text{CAD}.
\end{aligned}
\]

Hence, the fuzzy one-time development cost interval is:
\[
AI\_DevCost = [\,23{,}229.18,\; 36{,}499.52\,]~\text{CAD}, \qquad 
\text{with expected (defuzzified) value } 29{,}827.26~\text{CAD.}
\]

This total represents the full in-house human-resource expenditure to develop and validate the AI model, excluding compute or infrastructure charges (e.g., GPU depreciation, cloud instances) that were locally absorbed in Brugnara~\textit{et al.}.  

As noted earlier, due to limitations in obtaining public data needed for accurate cost estimation, the goal is not to achieve exact cost precision but to demonstrate how the framework structures and quantifies each component of AI development transparently.

\subsubsection{Recurring and Other Costs}
\label{sec:ai_recurring_costs}

These represent the ongoing costs of operating and maintaining the deployed AI solution. 
Because the model was developed and hosted in-house, no commercial software licensing or external vendor fees are included; only the recurring infrastructure and human-resource costs are captured.

In this case study, the model is assumed to be deployed on AWS SageMaker for managed inference and monitoring.  
The AWS Pricing Calculator\footnote{\url{https://calculator.aws/\# /createCalculator/SageMaker}}
 was used to estimate monthly infrastructure costs using the configuration shown in Table~\ref{tab:sagemaker_config}.  
Because GPU-equipped instances are unavailable in the Canada--West (Calgary) region, deployment was modeled in the Canada--Central region to access NVIDIA GPUs comparable to those used during model training 
(NVIDIA A100 (40 GB VRAM), using a DGX A100 system with 4 GPUs).

The total estimated monthly infrastructure cost is therefore:
\[
C_{\text{AWS}} = 8{,}050.86~\text{CAD/month}.
\]

\paragraph{Human resource component.}
As the AI solution is managed internally, we assume 1 ML engineer is responsible for model upkeep, periodic retraining, and system health monitoring.  
From Table~\ref{tab:ai_ccr_rates}, the ML engineer’s hourly rate is 44.10~CAD/hour.  
Assuming a standard 40-hour work week:
\[
H_{\text{month}} = 40 \times \frac{52}{12} = 173.33\ \text{hours/month},
\]
The monthly salary cost is:
\[
C_{\text{ML,month}} = 44.10 \times 173.33 = 7{,}644\ \text{CAD/month}.
\]

\paragraph{Monthly and annual recurring cost formulas.}
Since the model is in-house and incurs no additional licensing or SaaS charges beyond cloud deployment, we set \(C_{SaaS,t,m}=0\) and \(C_{other,t,m}=0\).  
The monthly operating cost is therefore:
\[
AI\_Cost_{t,m} = 8{,}050.86 + 7{,}644 = \mathbf{15{,}694.86\ CAD/month}.
\]
The annual cost in year \(t\) is:
\[
AI\_{Recurring}^{(t)} = 12 \times 15{,}694.86 = \mathbf{188{,}338.32\ CAD/year}.
\]

\begin{table}[h]
\caption{Monthly and annual recurring cost components for the LVO AI solution (production phase).}
\label{tab:recurring_costs}
\renewcommand{\arraystretch}{1.2}
\small
\begin{tabular*}{\textwidth}{@{\extracolsep\fill}p{0.7\textwidth}r@{}}
\toprule
\textbf{Component} & \textbf{Estimated Cost (CAD)} \\
\midrule
\multicolumn{2}{@{}l}{\textit{Monthly costs}} \\
AWS SageMaker infrastructure (\(C_{\text{AWS}}\)) & 8{,}050.86 \\
ML Engineer (40~hr/week @ 44.10~CAD/hr)           & 7{,}644.00 \\
\midrule
\textbf{Total monthly recurring cost}             & \textbf{15{,}694.86} \\
\midrule
\multicolumn{2}{@{}l}{\textit{Annualized cost (12 months)}} \\
AWS SageMaker (12 months)                         & 96{,}610.32 \\
ML Engineer (12 months)                           & 91{,}728.00 \\
\midrule
\textbf{Total annual recurring cost}              & \textbf{188{,}338.32} \\
\bottomrule
\end{tabular*}
\end{table}

\subsection{Cost of Service with AI-Assisted LVO Detection}
\label{sec:ct_service_with_ai}

The CT suite is a key component of the acute stroke imaging pathway, where non-contrast CT (NCCT) and CT angiography (CTA) are performed to identify large vessel occlusion (LVO). 
In the conventional workflow, both scans are manually interpreted by the radiologist and stroke neurologist before results are communicated to the intervention team. 
With the integration of the AI model trained for LVO detection, the overall CT-suite structure remains unchanged—\textbf{only the interpretation step is modified}. 

\begin{figure}[htbp]
  \centering
  \includegraphics[width=0.9\textwidth]{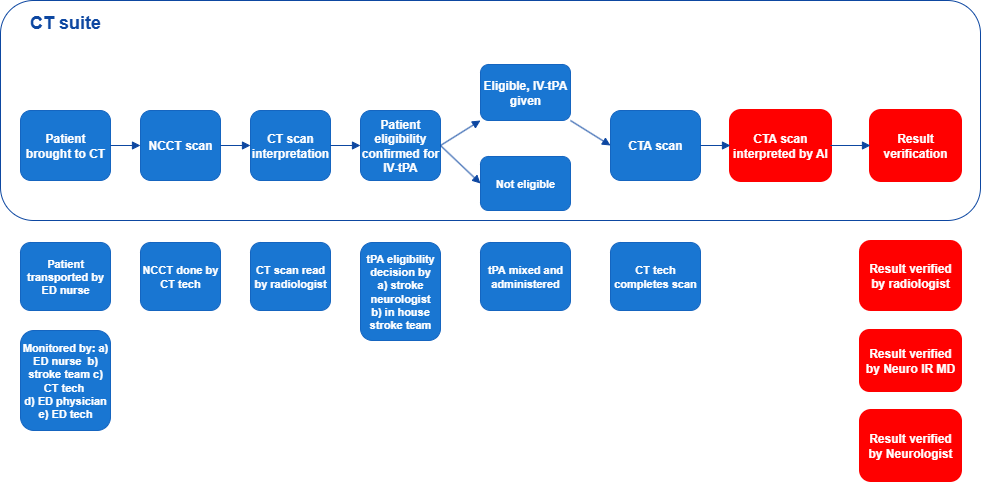}
  \caption{Integration of AI-assisted CTA interpretation within the CT suite.
  The AI replaces the manual interpretation step, while expert verification remains unchanged.}
  \Description{Block diagram of the CT stroke pathway showing where the AI-assisted CTA interpretation module is inserted between image acquisition and expert verification, while all other workflow steps remain unchanged.}
  \label{fig:ct_ai_suite_overview}
\end{figure}

The AI automatically analyzes CTA images to identify potential LVOs and generates a structured visual output highlighting the affected vessel. 
This output is then verified by the radiologist, neurointerventional physician, and stroke neurologist before confirmation and communication to the treatment team. 
All preceding activities, including patient positioning, image acquisition, and data transfer, remain identical to the standard process (Figure~\ref{fig:ct_pathway}). 
Figure~\ref{fig:ct_ai_suite_overview} illustrates how the AI system is integrated within the CT-suite workflow.

\noindent
The costing for this section, therefore, reflects the same activities and resources as the conventional CT suite, with the only adjustment applied to the CTA interpretation step where AI is introduced. 
The \textbf{AI inference time} is not charged within the CT-suite service cost because it has already been accounted for under the AI development and recurring operational cost components (Sections~\ref{sec:ai_cost_model}--\ref{sec:ai_recurring_costs}). 
This ensures that the computational expense associated with model execution is captured once, avoiding any double-counting of costs across the workflow. 
Human verification times are included using fuzzy estimates, as summarized in Table~\ref{tab:ct_suite_with_ai}.

Aggregating all pathway sections, including the CT-suite adjustments presented in Table~\ref{tab:ct_suite_with_ai}, yields the following cost per patient (Expected, Low, High):

\[
\text{Total Cost per patient} = \big[\,\mathbf{12{,}044.80},\; 11{,}466.37,\; 12{,}913.72\,\big]\ \text{CAD}.
\]

Assuming an annual patient volume of 5{,}000:

\[
C_{\text{annual}}^{\text{expected}} = 60{,}224{,}000 \text{ CAD}, \quad
C_{\text{annual}}^{\text{low}} = 57{,}331{,}850 \text{ CAD}, \quad
C_{\text{annual}}^{\text{high}} = 64{,}568{,}600 \text{ CAD.}
\]

\noindent
These figures represent the total service delivery cost for the stroke pathway when AI-assisted LVO detection is incorporated in the CT suite.

\subsection{ROI and Economic Performance Metrics}
\label{sec:roi_metrics}

In this subsection, we translate the data from the previous sections into clear, interpretable financial indicators.  
While previous sections estimated the cost of conventional and AI-enhanced stroke pathways, this section integrates all components—development, recurring, and service delivery costs—to estimate the ROI of adopting the AI-assisted LVO detection solution.  
The purpose is not only to determine whether the AI solution saves money but also to understand \emph{when} it begins to pay for itself and what operational factors influence that outcome.

In practice, health administrators and policymakers use these indicators to make informed decisions on early AI adoption.  
By quantifying when savings offset investment (break-even) and how efficiency changes over time (ROI trends), we enable proactive rather than retrospective evaluation.

\textit{Convention for clarity.} Unless otherwise stated, all results in this section are reported as \emph{expected (defuzzified)} values. Low and high fuzzy bounds for any term can be substituted into the same expressions to obtain corresponding ranges (e.g., point estimates become intervals)

\subsubsection*{Savings and Benefits}

The first step in assessing economic performance is to compute the savings per patient, which represents the cost reduction achieved when the AI system assists in CTA interpretation compared to the conventional workflow:

\[
\text{Savings per patient} =
\text{Cost}_{\text{without AI}} -
\text{Cost}_{\text{with AI}}.
\]

Using expected values:

\[
\begin{aligned}
\text{Cost}_{\text{without AI}} &= 12{,}099.45~\text{CAD},\\
\text{Cost}_{\text{with AI}} &= 12{,}044.80~\text{CAD},\\
\text{Savings per patient} &= 12{,}099.45 - 12{,}044.80 = 54.65~\text{CAD}.
\end{aligned}
\]

This means that, on average, each patient treated under the AI-assisted pathway costs about \$54.65 less than under the conventional pathway.  
Although this per-patient gain appears small, it scales with the number of patients treated annually.

\subsubsection*{Annual Benefit (Service-Side Savings)}

The \textbf{annual benefit} measures the total annual savings generated from applying the AI solution to the hospital’s full patient volume.  
It scales the per-patient savings by the total number of stroke patients treated each year:

\[
\text{Annual Benefit}_t
=
\text{Annual Cost}^{(\text{no AI})}_t
-
\text{Annual Cost}^{(\text{with AI})}_t.
\]

\[
\begin{aligned}
\text{Annual Cost}^{(\text{no AI})}&=60{,}497{,}250,\\
\text{Annual Cost}^{(\text{with AI})}&=60{,}224{,}000,\\
\Rightarrow\ \text{Annual Benefit}_{\text{(expected)}}&=60{,}497{,}250 - 60{,}224{,}000=273{,}250~\text{CAD}.
\end{aligned}
\]

\emph{Note:} This quantity reflects \textbf{service efficiency only}; it does \emph{not} include AI development or recurring costs. Consequently, if volumes and service times stay stable, $\text{Annual Benefit}_t$ is similar in later years.

\subsubsection*{AI Cost Over Time}

To correctly compute net savings and ROI, it is important to distinguish between the one-time and recurring components of AI costs.  
In Year~1, both the \textbf{development cost} and the \textbf{recurring cost} are incurred, whereas from Year~2 onward, only the recurring cost applies:

\[
\text{AI Cost}_t = 
\begin{cases}
\text{AI}_{\text{Dev}} + \text{AI}_{\text{Recurring}}, & t = 1, \\[4pt]
\text{AI}_{\text{Recurring}}, & t \ge 2.
\end{cases}
\]

Using expected values:

\[
\begin{aligned}
\text{AI}_{\text{Dev}} &= 29{,}827.26~\text{CAD},\\
\text{AI}_{\text{Recurring}} &= 188{,}338.32~\text{CAD},\\
\Rightarrow\ \text{AI Cost}_{\text{Year 1}} &= 218{,}165.58~\text{CAD},\\
\Rightarrow\ \text{AI Cost}_{t \ge 2} &= 188{,}338.32~\text{CAD}.
\end{aligned}
\]

Thus, the overall cost burden is highest in the first year and stabilizes thereafter.

\subsubsection*{Net Savings}

The \textbf{net savings} represent the true financial result after accounting for both service-level benefits and AI-related expenses:

\[
\text{Net Savings}_t =
\text{Annual Benefit}_t -
\text{AI Cost}_t.
\]

Substituting expected values:

\[
\begin{aligned}
\text{Net Savings}_{\text{Year 1}} &= 273{,}250 - 218{,}165.58 = 55{,}084.42~\text{CAD},\\[3pt]
\text{Net Savings}_{t \ge 2} &= 273{,}250 - 188{,}338.32 = 84{,}911.68~\text{CAD}.
\end{aligned}
\]

In Year~1, the net savings are lower due to the inclusion of the one-time development cost.  
From Year~2 onward, since only recurring costs are included, the deficit narrows—indicating gradual improvement in the financial trajectory.

\subsubsection*{Break-even Analysis}

The \textbf{break-even point} identifies when the cumulative savings from using AI equal the total investment made to build and operate it.  
At this point, the system ceases to run at a cumulative loss; all subsequent savings contribute to net profit:

\begin{equation}
\text{Patients to break even} =
\frac{\text{AI system cost}}{\text{Savings per patient}}.
\end{equation}

The expected AI system cost used in this calculation is the sum of the expected one-time development cost and the expected annual recurring cost.  
Substituting this value along with the expected per-patient savings yields:

\[
\begin{aligned}
\text{Patients to break even (expected)} &=
\frac{218{,}165.58}{54.65}
= 3{,}992.05~\text{patients}.
\end{aligned}
\]

Assuming an annual stroke volume of \(5{,}000\) patients, this corresponds to:

\[
\text{Months to break even (expected)} =
\frac{3{,}992.05}{5{,}000} \times 12
= 9.58~\text{months}.
\]

To account for uncertainty, the same calculation can be repeated using the \emph{low} and \emph{high} fuzzy bounds of the AI system cost:  
\[
AI_{\text{system,low}} = 211{,}567.50~\text{CAD}, \quad
AI_{\text{system,high}} = 224{,}837.80~\text{CAD}.
\]
This produces a corresponding break-even interval:

\[
\begin{aligned}
\text{Patients to break even (low bound)} &= \frac{211{,}567.50}{54.65} = 3{,}871.31~\text{patients},\\
\text{Patients to break even (high bound)} &= \frac{224{,}837.84}{54.65} = 4{,}114.14~\text{patients}.
\end{aligned}
\]

Converting to time under a constant 5,000-patient annual volume gives a range of:

\[
\text{Time to break even} \in [\,9.29~\text{months},9.87~\text{months}\,],
\]

Thus, under expected conditions, the hospital would need to process about 3,992 stroke cases—or operate for 9.58 months—before cumulative savings offset the first year AI investment.  
 
It should also be noted that yearly net savings may be negative if recurring costs exceed the annual benefit.

Likewise, we can compute the expected time to break even on the \emph{development} cost alone, i.e, the costs to develop the AI solution. Using the expected per-patient savings and annual volume:

\[
\text{Patients to break even (dev-only)} =
\frac{AI\_DevCost_{\text{expected}}}{\text{Savings per patient}} =
\frac{29{,}827.26}{54.65} =
545.78~\text{patients}.
\]

At an annual volume of \(5{,}000\) patients, the corresponding time is:

\[
\text{Time to break even (dev-only)} =
\frac{545.78}{5{,}000}\times 12 \approx
1.31~\text{months}\;(\approx 5.7~\text{weeks}).
\]

Under expected assumptions, the one-time development investment 
is recovered in approximately \(1.31\)~months. 
This calculation excludes recurring (operational) costs and reflects only service-side savings, paying down the development outlay.

\subsubsection*{Return on Investment (ROI)}

\[
ROI_t =
\frac{\text{Annual Benefit}_t - \text{AI Cost}_t}
     {\text{AI Cost}_t}
\times 100\%.
\]

Substituting expected values:

\[
\begin{aligned}
ROI_{\text{Year 1}} &=
\frac{273{,}250 - 218{,}165.58}{218{,}165.58} \times 100
= 25.25\%,\\[4pt]
ROI_{t \ge 2} &=
\frac{273{,}250 - 188{,}338.32}{188{,}338.32} \times 100
= 45.08\%.
\end{aligned}
\]

This means that in Year~1, every dollar invested in AI yields a net profit of about 25 cents; in subsequent years, the net profit improves to roughly 45 cents per dollar as development costs are excluded.  

\subsubsection*{Cumulative ROI over Five Years}

Cumulative ROI captures the combined financial outcome across multiple years:

\[
ROI_{\text{cum},T} =
\frac{\sum_{t=1}^{T} \text{Annual Benefit}_t -
      \sum_{t=1}^{T} \text{AI Cost}_t}
     {\sum_{t=1}^{T} \text{AI Cost}_t}
\times 100\%.
\]

Substituting expected yearly values gives:

\[
\begin{aligned}
ROI_{\text{cum},1} &= 25.25\%,\\
ROI_{\text{cum},2} &= 41.78\%,\\
ROI_{\text{cum},3} &= 42.82\%,\\
ROI_{\text{cum},4} &= 43.36\%,\\
ROI_{\text{cum},5} &= 43.70\%.
\end{aligned}
\]

The cumulative ROI thus improves gradually as development costs are amortized over multiple years.  

\begin{figure}[htbp]
  \centering
  \includegraphics[width=0.85\textwidth]{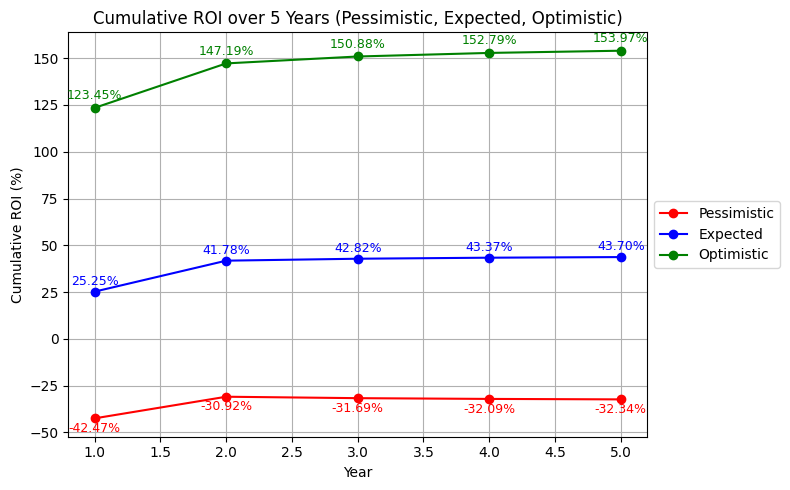}
  \caption{Cumulative ROI over five years under pessimistic, expected, and optimistic scenarios as one-time development costs are amortized. The expected case demonstrates steady positive returns, reaching approximately 44\% by Year 5. The pessimistic scenario remains negative throughout, while the optimistic scenario shows substantial gains exceeding 150\% by Year 5.}
  \Description{Line chart with years one through five on the horizontal axis and cumulative ROI as a percentage on the vertical axis. Three lines are plotted. The optimistic line rises steeply to above 150 percent by Year 5, the expected line rises from about 25 percent in Year 1 to roughly 44 percent by Year 5, and the pessimistic line stays below zero across all five years.}
  \label{fig:cumulative_roi}
\end{figure}

\subsection{Sensitivity Analysis}
\label{sec:sensitivity}

To understand how patient throughput affects the economic viability of adopting the AI solution, a one-way sensitivity analysis was performed. In this analysis, the Year~1 AI costs (which include one-time development and first-year recurring operating costs) were held constant, while the annual patient volume was varied from 1,000 to 10,000 cases per year. Year~1 is emphasized because it reflects the combined effect of the upfront investment and initial operating costs before any cost reductions in subsequent years. Importantly, the \textbf{AI cost used in this analysis is the \emph{expected} (defuzzified) AI cost}, rather than the low or high fuzzy bounds. The \textbf{expected} case therefore represents the central estimate of cost and benefit values, while the \textbf{pessimistic} and \textbf{optimistic} cases correspond to the lower and upper fuzzy bounds, respectively. Under this expected-cost assumption, the analysis indicates a \textbf{break-even point of approximately 3{,}992 patients per year}.

\subsubsection{Effect of Patient Volume on ROI}

The results show that the economic value of the AI solution depends strongly on annual patient volume. At low volumes (below approximately 4,000 cases per year), the annual savings generated by faster workflow and reduced interpretation time are not enough to recover the upfront investment, resulting in negative net savings and negative ROI in the expected case. As volume increases, the fixed investment cost is distributed across more patients, while per-patient savings accumulate. This leads to a steady increase in ROI with scale.

The break-even point in the expected case occurs at approximately 3,992 patients. At this volume, the cumulative savings equal the Year~1 cost, and ROI becomes positive. Beyond this point, ROI continues to rise. At the commonly observed provincial volume of 5,000 stroke patients per year, the expected case yields a positive ROI (25.25\%) and a positive annual net savings, indicating that the investment is economically viable under typical operating conditions.

\begin{figure}[htbp]
\centering
\includegraphics[width=0.85\textwidth]{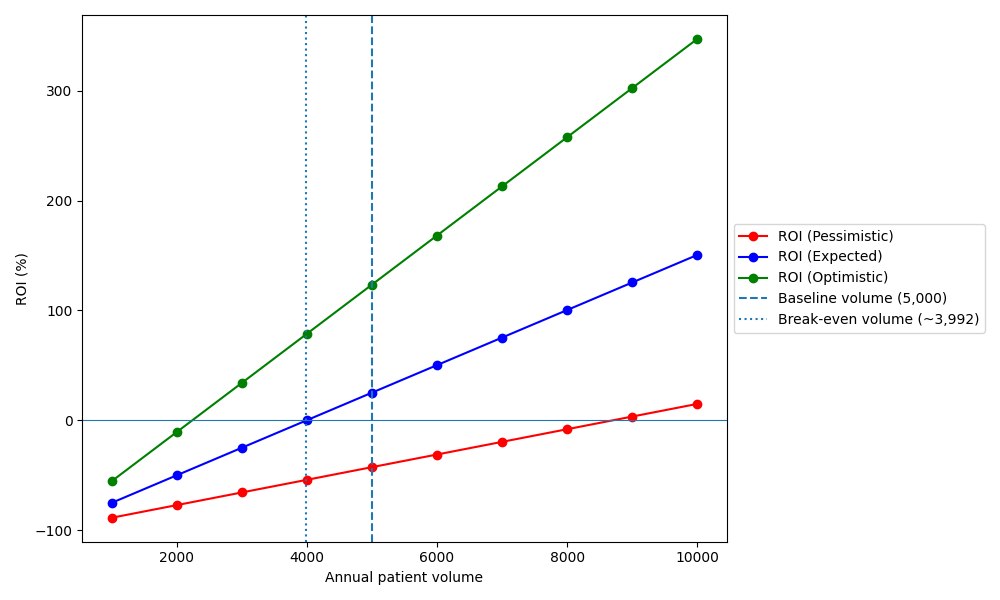}
\caption{ROI vs. annual patient volume across pessimistic, expected, and optimistic scenarios.}
\Description{Line chart with annual patient volume from 1,000 to 10,000 cases on the horizontal axis and Year 1 ROI as a percentage on the vertical axis. Three lines are plotted for the pessimistic, expected, and optimistic scenarios; all rise as volume increases. The expected line crosses zero ROI at approximately 3,992 patients per year, the optimistic line crosses zero near 3,000 patients, and the pessimistic line crosses at a higher volume.}
\label{fig:roi_vs_volume}
\end{figure}

Differences between the pessimistic and optimistic cases highlight the role of implementation efficiency. In the pessimistic scenario, break-even occurs later because workflow gains are smaller and verification time remains higher. In contrast, under optimistic conditions, the AI system becomes profitable at volumes near or slightly below 3,000 patients per year. These relationships are illustrated in Figure~\ref{fig:roi_vs_volume}, which shows how ROI increases with annual patient volume under pessimistic, expected, and optimistic scenarios.

\subsubsection{Time to Break-Even}

The time required to recover the initial investment decreases as patient volume increases. At 4,000 patients per year, the expected case reaches break-even in approximately 12 months. At 5,000 to 6,000 patients per year, break-even is reached in roughly 8--10 months. This indicates that once sufficient patient volume is reached, the AI solution begins generating net financial benefit within the first year of use.

\section{Limitations}

The FLARE framework, while offering a structured and uncertainty-aware method for evaluating the economic impact of AI in healthcare, has several limitations.  
First, FLARE relies on FL--TDABC, which requires detailed activity maps, personnel rates, and time estimates. Such inputs are often unavailable in published studies, resulting in the use of expert consultation or approximate values. Although fuzzy logic helps represent uncertainty in activity durations, the reliability of the results still depends on the quality and accuracy of these underlying assumptions.

In addition, while a time–motion study was conducted with radiologists to verify clinical activities and time estimates used in the case study, FLARE itself has not been formally evaluated through a structured user study. The framework was not assessed by end users such as hospital administrators, health economists, or policymakers in terms of usability, interpretability, or decision-making impact. Nevertheless, the costing logic underlying FLARE is based on TDABC and fuzzy logic, both of which are well established and widely validated in prior academic and industry work \cite{keel2017time}\cite{koster2023dealing}. Accordingly, this limitation relates to user-facing evaluation rather than to the validity of the underlying economic methodology.

Second, FLARE models uncertainty primarily in activity durations and cost parameters using triangular fuzzy numbers. Other sources of uncertainty are not explicitly represented. As a result, the framework may underestimate the full range of real-world uncertainty in some deployment scenarios.

Third, the framework does not explicitly model macroeconomic factors such as inflation, interest rates, or the time value of money through formal discounting. All costs and benefits are expressed in constant monetary terms, and ROI is computed without applying a discount rate. This design choice reflects the intended use of FLARE as an early-stage decision-support tool, where reliable assumptions about long-term discount rates and inflation trajectories are often unavailable.

However, FLARE’s flexible parameterization partially mitigates this limitation. Because the framework allows users to specify costs at monthly or yearly resolution, inflationary effects can be incorporated indirectly by adjusting input parameters over time. For example, users may increase recurring operational costs, personnel rates, or development-related resource costs in later periods to reflect anticipated inflation or salary growth. In this way, while inflation is not modeled explicitly as a financial parameter, its effects can be explored through scenario-based adjustments to input values.

Fourth, the case study focuses on a single stroke-imaging workflow in a specific regional context (Calgary, Alberta). Personnel rates, material costs, and workflow characteristics vary across institutions and health systems. While the FLARE framework itself is generalizable, the numerical results presented in this paper should not be interpreted as universally applicable without local recalibration of parameters.

Finally, the framework has not yet been validated against detailed ground-truth costing datasets, nor has it been directly compared with alternative costing methodologies such as micro-costing or top-down costing approaches. This limitation reflects the lack of publicly available, high-resolution cost datasets suitable for benchmarking. As a result, the current evaluation focuses on demonstrating the structure, transparency, and uncertainty-aware capabilities of FLARE rather than assessing its accuracy in reproducing true institutional costs.

\section{Discussion}

The results of this analysis show that the economic viability of adopting the AI solution for LVO detection is closely linked to patient volume and workflow efficiency. By using FL--TDABC to model the clinical activities involved in the CT stroke pathway, FLARE makes the relationship between resource use, time, and cost transparent. This allows us to see not only where costs are incurred, but also where financial savings are generated through reduced interpretation time and streamlined workflow.

The sensitivity results indicate that when patient volume is below approximately 4,000 cases per year, the annual savings are not sufficient to offset the Year~1 investment, resulting in negative ROI. In such cases, the AI solution does not become cost-saving in the first year. However, this does not imply that the AI system is unsuitable. Rather, it highlights that the economic benefit largely depends on how the system is implemented and scaled. For instance, reducing development costs by streamlining annotation workflows, minimizing personnel time, or pursuing collaborative model development, as well as adopting subscription-based AI services instead of local in-house development, can substantially lower the initial investment. Likewise, selecting optimized cloud instance configurations or shared GPU resources can reduce ongoing server expenses. These adjustments can move the break-even point to lower patient volumes.

As patient volume increases, the fixed development cost is spread across more cases, while per-patient savings accumulate. This causes ROI to rise steadily. At common stroke volumes of approximately 5,000 cases per year, the investment yields a positive return within the first year. This shows that the AI solution is financially viable in hospitals or regions with moderate to high patient throughput. In contrast, smaller centers may require alternative strategies—such as regional image routing, centralized radiology reading models, or phased deployment—to achieve similar value.

There is also meaningful variation between pessimistic and optimistic cases. In the pessimistic scenario, lower workflow efficiency and longer verification time delay cost recovery, whereas in the optimistic scenario, improved workflow integration leads to positive ROI at lower volumes. This highlights that financial outcomes are influenced not only by model performance but also by how effectively the AI system is integrated into clinical practice. Successful adoption, therefore, requires attention to workflow redesign, clinician trust, training, and usability, in addition to technical deployment.

Finally, while this sensitivity analysis varied only annual patient volume to maintain clarity in interpretation, the FLARE framework allows any model input to be varied to examine overall economic impact. For example, the AI development cost can be adjusted by changing personnel wages, labeling effort, or GPU/server usage. Similarly, verification time, cloud hosting strategy, and staffing configuration can be modified to reflect different implementation settings. Varying these parameters provides a more comprehensive understanding of how costs and benefits shift under different real-world conditions, enabling stakeholders to evaluate a wider range of operational strategies and deployment models.

\section{Conclusion}

This paper introduced FLARE, a Fuzzy-Logic and Time-Driven Activity-Based Return on Investment Evaluation framework designed to support evidence-based decision-making for AI adoption in healthcare. FLARE was proposed to address a key gap in current evaluation practices, where most studies emphasize performance metrics while overlooking the financial and operational implications of integrating AI systems into real clinical workflows. By combining FL--TDABC costing with fuzzy logic and ROI analysis, FLARE provides a transparent, activity-based, and uncertainty-aware method for estimating both the cost of delivering care and the economic impact of introducing AI technologies.

The EHTA in acute ischemic stroke imaging demonstrated how FLARE can be applied to a real-world clinical pathway. The results showed that the economic value of adopting AI depends on both workflow efficiency and patient volume. Using the expected (defuzzified) AI cost, the break-even point was estimated at approximately 3{,}992 patients per year. At typical stroke volumes of around 5{,}000 patients per year, the AI solution produced a positive return on investment within the first year of deployment. More broadly, the findings highlight that realizing economic benefit is not guaranteed by algorithmic performance alone but depends on effectively aligning development costs, personnel effort, infrastructure decisions, and clinical workflow integration.

FLARE makes these trade-offs explicit, providing stakeholders with a structured way to identify when and where AI adoption is economically viable, and where adjustments—such as lowering development costs, optimizing verification workflows, or scaling deployment across sites—can improve financial outcomes.

Looking forward, the next stage involves translating the framework into an interactive software platform that allows users to adjust assumptions, model uncertainty, and visualize ROI outcomes in real time. This step transforms FLARE from an analytical framework into an operational decision-support tool for clinicians, administrators, and policymakers.

\begin{acks}
\end{acks}

\bibliographystyle{ACM-Reference-Format}
\bibliography{ref.bib}

\appendix

\section{Supplementary Materials for CT Pathway Cost Modeling}
\label{appendix:ct_pathway_tables}
\begin{table}[htpb]
\centering
\caption{Personnel hourly rates and capacity cost rates (CCR) used for FL--TDABC modeling. All rates are estimates for a Canadian academic hospital setting..}
\renewcommand{\arraystretch}{1.3}
\small
\begin{tabular}{lcc}
\toprule
\textbf{Resource (Personnel)} & \textbf{Hourly Rate (CAD)} & \textbf{CCR (CAD/min)} \\
\midrule
ED clerk & 24.50 & 0.41 \\
ED technician & 44.08 & 0.73 \\
ED physician & 130.24 & 2.17 \\
ED nurse & 47.38 & 0.79 \\
NIR nurse & 58.51 & 0.98 \\
NIR technologist & 44.61 & 0.74 \\
Interventionalist physician & 130.24 & 2.17 \\
NIR PA or NP & 44.61 & 0.74 \\
Anesthesiologist & 82.72 & 1.38 \\
Stroke Resident & 30.20 & 0.50 \\
Stroke Neurologist & 44.61 & 0.74 \\
CT technologist & 44.61 & 0.74 \\
CT nurse & 58.51 & 0.98 \\
Radiologist & 44.61 & 0.74 \\
\bottomrule
\end{tabular}
\label{tab:personnel_ccr}
\end{table}

\begin{table}[h]
\caption{Material cost estimates used in the CT pathway cost model (in CAD).}\label{tab:materials_cost}
\renewcommand{\arraystretch}{1.3}
\small
\begin{tabular}{@{}
>{\raggedright\arraybackslash}p{4.1cm}
>{\raggedright\arraybackslash}p{2.4cm}
>{\raggedright\arraybackslash}p{4.8cm}
@{}}
\toprule
\textbf{Material} & \textbf{Unit Cost (CAD)} & \textbf{Source} \\
\midrule
Lab tests & 0 & -- \\
Contrast (CTA) & 125 & Mayfair Diagnostics \\
IV--tPA (Alteplase 100 mg) & 2,746 & Federal Patented Medicine Prices Review Board (PMPRB) \\
Stroke pack & 5,500 & -- \\
\bottomrule
\end{tabular}
\end{table}

\setlength\LTleft{0pt}
\setlength\LTright{0pt}
\setlength{\tabcolsep}{2pt}

\begin{landscape}
\footnotesize
\setlength{\LTcapwidth}{\linewidth}
\setlength\LTleft{0pt}
\setlength\LTright{0pt}
\setlength{\tabcolsep}{2pt}

\begin{longtable}{@{}>{\raggedright\arraybackslash}p{0.098\linewidth}>{\raggedright\arraybackslash}p{0.140\linewidth}>{\raggedright\arraybackslash}p{0.140\linewidth}>{\raggedright\arraybackslash}p{0.098\linewidth}>{\raggedright\arraybackslash}p{0.082\linewidth}>{\centering\arraybackslash}p{0.025\linewidth}>{\centering\arraybackslash}p{0.025\linewidth}>{\centering\arraybackslash}p{0.025\linewidth}>{\centering\arraybackslash}p{0.066\linewidth}>{\centering\arraybackslash}p{0.072\linewidth}>{\centering\arraybackslash}p{0.052\linewidth}>{\centering\arraybackslash}p{0.052\linewidth}@{}}
\caption{Process map of the conventional CT pathway. The activity structure and workflow steps were obtained from Sangha et al. \cite{sangha2024time}, while all time estimates and cost parameters were obtained through a time–motion study conducted with a practicing radiologist in a tertiary academic stroke centre.}

\label{tab:stroke_costing}\\

\toprule
\textbf{Activity} & \textbf{Sub-tasks} & \textbf{Personnel} & \textbf{Equipment} & \textbf{Material} &
\textbf{a} & \textbf{b} & \textbf{c} & \textbf{Expected Time (min)} &
\textbf{Expected Cost (CAD)} & \textbf{Cost Low} & \textbf{Cost High} \\
\midrule
\endfirsthead

\multicolumn{12}{l}{\footnotesize\emph{(Continued from previous page)}}\\
\toprule
\textbf{Activity} & \textbf{Sub-tasks} & \textbf{Personnel} & \textbf{Equipment} & \textbf{Material} &
\textbf{a} & \textbf{b} & \textbf{c} & \textbf{Expected Time (min)} &
\textbf{Expected Cost (CAD)} & \textbf{Cost Low} & \textbf{Cost High} \\
\midrule
\endhead

\midrule
\multicolumn{12}{r}{\footnotesize\emph{(Continued on next page)}}\\
\endfoot

\bottomrule
\endlastfoot


\multicolumn{12}{c}{\cellcolor{gray!50}\textbf{ED Triage}} \\
\midrule
\multirow{3}{=}{ED assessment with positive stroke symptoms} 
& Placed on stretcher & ED clerk; ED nurse & Stretcher; Monitor & -- & 7 & 12 & 60 & 26.33 & 31.59 & 8.4 & 72 \\
\cmidrule(lr){2-12}
& Code stroke called by triage & ED nurse; ED physician & -- & -- & 5 & 10 & 60 & 25 & 74 & 14.8 & 177.6 \\
\midrule
\multirow{3}{=}{Patient monitoring} 
& Placed on stretcher & ED nurse; ED technician & Monitor; Stretcher & -- & 5 & 10 & 60 & 25 & 38 & 7.6 & 91.2 \\
\cmidrule(lr){2-12}
& Hooked up to monitor & & & & & & & & & & \\
\cmidrule(lr){2-12}
& Vitals and weight taken & & & & & & & & & & \\
\midrule
\multirow{4}{=}{ED assessment} 
& Patient medical and medication history taken & ED physician; ED nurse & Stretcher; Monitor & Lab tests & 10 & 30 & 90 & 43.33 & 128.25 & 29.6 & 266.4 \\
\cmidrule(lr){2-12}
& Placement of IV & & & & & & & & & & \\
\cmidrule(lr){2-12}
& Lab panel sent & & & & & & & & & & \\
\cmidrule(lr){2-12}
& Assessment of blood glucose & & & & & & & & & & \\
\midrule
\multirow{2}{=}{Stroke team evaluation} 
& NIHSS assessment & Stroke Resident; Stroke Neurologist; ED physician; ED nurse & Monitor; Stretcher & -- & 15 & 30 & 90 & 45 & 189 & 63 & 378 \\
\cmidrule(lr){2-12}
& tPA prepared & & & & & & & & & & \\
\midrule
\multicolumn{12}{c}{\cellcolor{gray!50}\textbf{CT Suite}} \\
\midrule
\multirow{3}{=}{Patient brought to CT} 
& Patient transported by ED nurse & ED nurse; CT nurse & Stretcher; Monitor & -- & 10 & 15 & 30 & 18.33 & 32.44 & 17.7 & 53.1 \\
\cmidrule(lr){2-12}
& Monitored by: a) ED nurse, b) stroke team c) CT tech & & & & & & & & & & \\
\cmidrule(lr){2-12}
& d) ED physician e) ED tech & & & & & & & & & & \\
\midrule
NCCT scan & NCCT done by CT tech & CT technologist; CT nurse & CTH (CT suite only); Monitor; Stretcher & -- & 4 & 6 & 8 & 6 & 10.32 & 6.88 & 13.76 \\
\midrule
CT scan interpretation & CT scan read by radiologist & Radiologist; Stroke Neurologist & -- & -- & 5 & 7 & 15 & 9 & 13.32 & 7.4 & 13.76 \\
\midrule
\multirow{2}{=}{Patient eligibility confirmed for IV-tPA} 
& tPA eligibility decision by: a) stroke neurologist & Stroke Neurologist; Stroke Resident; ED physician; ED nurse & IV-tPA (Alteplase 100 mg) & -- & 5 & 10 & 30 & 15 & 2809 & 2767 & 2872 \\
\cmidrule(lr){2-12}
& b) in house stroke team & & & & & & & & & & \\
\midrule
Eligible, IV-tPA given / Not eligible & tPA mixed and administered & & & & & & & & & & \\
\midrule
CTA scan & CT tech completes scan & CT technologist; CT nurse & CTH (CT suite only); Monitor; Stretcher & Contrast (CTA) & 2 & 4 & 10 & 5.33 & 134.1676 & 128.44 & 142.2 \\
\midrule
\multirow{3}{=}{CTA scan interpreted} 
& CTA scan interpreted by radiologist & Radiologist; Stroke Neurologist & -- & -- & 5 & 10 & 30 & 15 & 22.2 & 7.4 & 44.4 \\
\cmidrule(lr){2-12}
& CTA scan interpreted by Neuro IR MD & & & & & & & & & & \\
\cmidrule(lr){2-12}
& CTA scan interpreted by Neurologist & & & & & & & & & & \\
\midrule
LVO / No LVO & LVO confirmation & Radiologist; Stroke Neurologist & -- & -- & 5 & 8 & 15 & 9.33 & 13.80 & 7.4 & 22.2 \\
\midrule
\multicolumn{12}{c}{\cellcolor{gray!50}\textbf{NIR Suite}} \\
\midrule
\multirow{3}{=}{Patient confirmed candidate for LVO} 
& If LVO, decision for treatment made & Stroke Neurologist; ED physician; ED nurse & -- & -- & 10 & 12 & 30 & 17.33 & 64.12 & 37 & 111 \\
\cmidrule(lr){2-12}
& If treatment, on call NIR team called & & & & & & & & & & \\
\midrule
\multirow{7}{=}{On call NIR team called} 
& Patient transport to NIR suite by: a) Stroke neurophysician & NIR nurse; NIR technologist; Interventional Specialist physician; NIR PA or NP; Anesthesiologist & Biplane angiography; Monitor; Stretcher; pack & IV-tPA (Alteplase 100 mg); Stroke pack & 15 & 25 & 45 & 28.33 & 170.26 & 90.15 & 270.45 \\
\cmidrule(lr){2-12}
& b) ED nurse & & & & & & & & & & \\
\cmidrule(lr){2-12}
& c) ED nurse & & & & & & & & & & \\
\cmidrule(lr){2-12}
& IR suite staff called, includes: a) NIR nurse & & & & & & & & & & \\
\cmidrule(lr){2-12}
& b) NIR tech & & & & & & & & & & \\
\cmidrule(lr){2-12}
& c) Interventionalist MD & & & & & & & & & & \\
\cmidrule(lr){2-12}
& d) NIRN PA or NP & & & & & & & & & & \\
\cmidrule(lr){2-12}
& e) Anesthesiologist & & & & & & & & & & \\
\midrule
Groin puncture and EVT & Patient EVT by NIR suite staff & Interventionalist physician; NIR nurse; NIR technologist; Anesthesiologist & Biplane angiography; Monitor; Stretcher & IV-tPA (Alteplase 100 mg); Stroke pack & 10 & 15 & 45 & 23.33 & 8368.9491 & 8298.7 & 8483.15 \\
\midrule
\midrule
\multicolumn{5}{l}{\textbf{Total Cost per patient}} & & & & & \textbf{12099.45} & \textbf{11491.47} & \textbf{13011.22} \\
\multicolumn{5}{l}{\textbf{Annual Patient Volume}} & & & & & \textbf{5000} & & \\
\multicolumn{5}{l}{\textbf{Annual Cost}} & & & & & \textbf{60497250} & \textbf{57457350} & \textbf{65056100} \\

\end{longtable}
\end{landscape}

\section{Step-by-Step FL--TDABC Application for the CT Stroke Pathway}
\label{appendix:ct_fltdabc_steps}

\paragraph{Step 1: Define the Service}

The service being analyzed is one full CT stroke pathway, defined from patient arrival in the ED with stroke symptoms to either confirmation or exclusion of LVO and, where eligible, treatment in the NIR suite.

\paragraph{Step 2: Define the Care Delivery Value Chain}

The CT pathway includes three main stages:
\begin{enumerate}
    \item \textbf{ED Triage} – initial assessment, monitoring, and activation of the stroke team.
    \item \textbf{CT Suite} – NCCT and CTA imaging, interpretation, and eligibility confirmation for thrombolysis or EVT.
    \item \textbf{NIR Suite} – preparation and performance of EVT for confirmed LVO cases.
\end{enumerate}

\paragraph{Step 3: Develop Process Maps}

Each stage is divided into micro-activities, specifying the personnel, equipment, and materials required.  
For example, in the CT suite, activities include:
\begin{itemize}
    \item NCCT scan acquisition by a CT technologist,
    \item CTA scan completion and interpretation,
    \item IV-tPA preparation (if indicated),
    \item Patient monitoring by ED and stroke team members.
\end{itemize}

\paragraph{Step 4: Obtain Fuzzy Time Estimates}

Each activity duration is represented as a triangular fuzzy number $(a,b,c)$ in minutes, corresponding to the minimum, most likely, and maximum time, respectively.  
The expected or “defuzzified” time is computed as:
\[
x^{*} = \frac{a + b + c}{3}.
\]

\paragraph{Step 5: Estimate Resource Costs}

For each personnel type, the hourly wage rate was converted to a minute-based rate (Capacity Cost Rate, CCR):
\[
CCR_i = \frac{\text{Hourly Rate}_i}{60}.
\]
For example, the CCR for a radiologist is
\[
CCR_{\text{Radiologist}} = \frac{44.61}{60} = 0.74 \text{ CAD/min}.
\]
Material costs (e.g., CTA contrast, IV-tPA, stroke pack) were obtained from Canadian sources such as Mayfair Diagnostics and the federal Patented Medicine Prices Review Board (PMPRB).

\paragraph{Step 6: Compute Activity Cost (Worked Example)}

The activity cost is calculated by multiplying the expected time ($x^{*}$) of each activity by the corresponding resource’s capacity cost rate (CCR). To account for uncertainty, the lower bound cost is obtained by substituting the minimum time estimate ($a$) for $x^{*}$, and the upper bound cost is obtained by substituting the maximum time estimate ($c$).  
This ensures that the resulting cost interval captures best- and worst-case workflow durations.

Example activity: “CTA scan interpreted.”

\begin{itemize}
    \item Personnel: Radiologist, Stroke Neurologist
    \item Fuzzy time estimate: $(a,b,c) = (5,7,9)$ minutes  
    \item Expected time:
    \[
    x^{*} = \frac{5 + 7 + 9}{3} = 7 \text{ minutes.}
    \]
    \item CCR (Radiologist) = 0.74 CAD/min
    \item CCR (Stroke Neurologist) = 0.74 CAD/min
\end{itemize}

\noindent Therefore, \text{expected cost per activity}
\[
\begin{aligned}
&= x^{*}\times CCR(\text{Radiologist})
 + x^{*}\times CCR(\text{Stroke Neurologist}) \\
&= 7\times 0.74 + 7\times 0.74 \\
&= 10.36\ \text{CAD}, \\[4pt]
\text{Low bound cost}
&= a\times CCR(\text{Radiologist})
 + a\times CCR(\text{Stroke Neurologist}) \\
&= 5\times 0.74 + 5\times 0.74 \\
&= 7.40\ \text{CAD}, \\[4pt]
\text{High bound cost}
&= c\times CCR(\text{Radiologist})
 + c\times CCR(\text{Stroke Neurologist}) \\
&= 9\times 0.74 + 9\times 0.74 \\
&= 13.32\ \text{CAD}.
\end{aligned}
\]

In the complete pathway costing, when multiple professionals contribute to the same activity, their individual costs are summed.  
Material costs (if any) are then added to obtain the total activity cost.

\paragraph{Step 7: Compute Total Cost per Patient}

Summing across all activities yields the per-patient expected, low, and high cost estimates:
\[
\text{Cost}_{\text{expected}} = 6259.72\text{ CAD}, \quad
\text{Cost}_{\text{low}} = 6065.22\text{ CAD}, \quad
\text{Cost}_{\text{high}} = 6464.22\text{ CAD}.
\]

\section{Supplementary Materials for AI Cost Modeling}
\label{appendix:ai_costing}

\subsection{Personnel Capacity Cost Rates (CCR)}

\begin{table}[htpb]
\centering
\caption{Personnel hourly rates and capacity cost rates (CCR) used for AI development costing.}
\renewcommand{\arraystretch}{1.2}\small
\begin{tabular}{lcc}
\toprule
\textbf{Personnel role} & \textbf{Hourly rate (CAD)} & \textbf{CCR (CAD/min)} \\
\midrule
PACS / IT Technician & 40.12 & 0.67 \\
ML Engineer & 44.10 & 0.74 \\
Data Scientist & 45.67 & 0.76 \\
Radiologist & 44.61 & 0.74 \\
\bottomrule
\end{tabular}
\label{tab:ai_ccr_rates}
\end{table}

\subsection{Full Process Map for the Heidelberg Training Cohort}

\begin{landscape}
\begin{table}[htpb]
\centering
\caption{Process map for the Heidelberg training cohort ($N{=}835$).}
\renewcommand{\arraystretch}{1.2}\small
\begin{tabular}{p{4.1cm} p{4.0cm} c c c c c c}
\toprule
\textbf{Activity} & \textbf{Personnel (CCR, CAD/min)} & \textbf{$a$ (min)} & \textbf{$b$ (min)} & \textbf{$c$ (min)} & \textbf{$x^{*}=\frac{a+b+c}{3}$} & \textbf{Expected Cost (CAD)} & \textbf{Low / High (CAD)} \\
\midrule
Data gathering & PACS/IT (0.67) & 3.00 & 4.00 & 5.00 & 4.00 & 2.68 & 2.01 / 3.35 \\
Data Preprocessing  & ML Engineer (0.735) & 1.12 & 1.38 & 1.88 & 1.46 & 1.07 & 0.82 / 1.38 \\
Data Labeling & Radiologist (0.74) & 17.00 & 22.00 & 27.00 & 22.00 & 16.28 & 12.58 / 19.98 \\
Model training \textsuperscript{$\dagger$} & ML Engineer (0.735) & 6.00 & 7.00 & 8.00 & 7 & 5.18 & 4.41 / 5.88 \\
\midrule
\textbf{Total per sample} & & & & & & \textbf{25.22} & \textbf{19.86 / 30.64} \\
\bottomrule
\end{tabular}
\label{tab:train_unit_costs}
\end{table}
\end{landscape}

\noindent\textit{Note.} \(\dagger\) Compute/GPU costs excluded (locally owned hardware as reported); line prices ML-engineer time only.

\subsection{Full Process Map for the Heidelberg Test Cohort}
\begin{landscape}
\begin{table}[htpb]
\centering
\caption{Process map for the Heidelberg internal test cohort ($N{=}344$).}
\renewcommand{\arraystretch}{1.2}\small
\begin{tabular}{p{4.1cm} p{4.0cm} c c c c c c}
\toprule
\textbf{Activity} & \textbf{Personnel (CCR, CAD/min)} & \textbf{$a$ (min)} & \textbf{$b$ (min)} & \textbf{$c$ (min)} & \textbf{$x^{*}=\frac{a+b+c}{3}$} & \textbf{Expected Cost (CAD)} & \textbf{Low / High (CAD)} \\
\midrule
Data gathering & PACS/IT (0.67) & 3.00 & 4.00 & 5.00 & 4.00 & 2.68 & 2.01 / 3.35 \\
Data Preprocessing & ML Engineer (0.735) & 1.12 & 1.38 & 1.88 & 1.46 & 1.08 & 0.83 / 1.39 \\
Data Labeling & Radiologist (0.74) & 17.00 & 22.00 & 27.00 & 22.00 & 16.28 & 12.58 / 19.98 \\
Inference  & ML Engineer (0.735) & 0.27 & 0.33 & 0.47 & 0.36 & 0.27 & 0.20 / 0.35 \\
Evaluation / review & Radiologist (0.74) & 5.00 & 7.00 & 9.00 & 7.00 & 5.18 & 3.70 / 6.66 \\
\midrule
\textbf{Total per sample (priced)} & & & & & & \textbf{25.49} & \textbf{19.32 / 31.73} \\
\bottomrule
\end{tabular}
\label{tab:test_unit_costs}
\end{table}
\end{landscape}

\begin{table}[htbp]
\caption{AWS SageMaker deployment configuration and estimated monthly cost (Canada--Central region).}\label{tab:sagemaker_config}
\renewcommand{\arraystretch}{1.2}
\small
\begin{tabular*}{\textwidth}{@{\extracolsep\fill}
>{\raggedright\arraybackslash}p{2.6cm}
>{\raggedright\arraybackslash}p{6.6cm}
>{\raggedright\arraybackslash}p{2.6cm}@{}}
\toprule
\textbf{Section} & \textbf{Parameter} & \textbf{Recommended / Selected Value} \\
\midrule
\textbf{Real-time inference instances} & Number of models deployed & 1 \\
& Number of models per endpoint & 1 \\
& Number of instances per endpoint & 1 \\
& Endpoint hours per day & 24 \\
& Endpoint days per month & 30 \\
& Selected instance type & \texttt{ml.g5.12xlarge} (GPU-enabled) \\
\midrule
\textbf{Model monitor} & Number of Model Monitor jobs per month & 30 \\
& Instances per Model Monitor job & 1 \\
& Hours per Model Monitor instance per job & 1 \\
& Selected instance type & \texttt{ml.c5.18xlarge} \\
\midrule
\textbf{ML storage} & Storage type & General Purpose SSD (gp2/gp3) \\
& Storage amount & 30~GB \\
\midrule
\textbf{Data processing} & Data processed IN & 50~GB/month (\(\sim\)500k images) \\
& Data processed OUT & 1~GB/month \\
\midrule
\textbf{Region} & AWS Region & Canada (Central) \\
\midrule
\textbf{Estimated monthly cost} & --- & \textbf{5,793.82~USD} (\(\sim\)\textbf{8,050.86~CAD}) \\
\bottomrule
\end{tabular*}
\end{table}

\subsection{Step-by-Step FL--TDABC Application for AI Development Cost}
\label{appendix:ai_dev_fltdabc_steps}

\paragraph{Step 1 (Service).} Unit is \emph{one data sample} processed through the development workflow.

\paragraph{Step 2 (Value chain).} Activities differ by cohort: training samples usually require gathering, preprocessing, labeling, inclusion in training, and (optionally) post-train checks; test samples generally require \emph{gathering, preprocessing, labeling/ground-truthing, automated inference, and evaluation/review}.

\paragraph{Step 3 (Process map).} Assign each activity to a primary role and its resource rate (Tables~\ref{tab:train_unit_costs}--\ref{tab:test_unit_costs}; see also Table~\ref{tab:ai_ccr_rates}).

\paragraph{Step 4 (Fuzzy times).} Each activity uses triangular fuzzy times \((a,b,c)\) (minutes) with expected time
\[
x^{*}=\frac{a+b+c}{3}.
\]

\paragraph{Step 5 (Estimate Resource Costs).} Hourly rates and CCR for resources are estimated (Table~\ref{tab:ai_ccr_rates}). 

\paragraph{Step 6 (Compute Activity Cost ).} 

Example activity: “Data Gathering”

\begin{itemize}
    \item Personnel: PACS / IT Technician
    \item Fuzzy time estimate: $(a,b,c) = (3,4,5)$ minutes  
    \item Expected time:
    \[
    x^{*} = \frac{3 + 4 + 5}{3} = 4 \text{ minutes.}
    \]
    \item CCR (PACS / IT Technician) = 0.67 CAD/min
\end{itemize}

\noindent Therefore, expected cost per activity
\[
\begin{aligned}
 &= x^{*} \times CCR(\text{PACS / IT Technician}) = 4 \times 0.67 = 2.68 \text{ CAD}, \\
\text{Low bound cost} &= a^{*} \times CCR(\text{PACS / IT Technician}) = 3 \times 0.67 = 2.01 \text{ CAD}, \\
\text{High bound cost} &= c^{*} \times CCR(\text{PACS / IT Technician}) = 5 \times 0.67 = 3.35 \text{ CAD}
\end{aligned}
\]

\paragraph{Step 7 (Compute Total Cost per Sample).} Summing across all activities yields the per-sample expected, low, and high cost estimates (Table~\ref{tab:train_unit_costs} and Table~\ref{tab:test_unit_costs}).

\subsection{Supplementary Tables for Service and AI Cost Modeling}
\label{appendix:service_ai_costing}

\subsection{CT Suite Process Map with AI-Assisted LVO Detection}
\label{appendix:ct_ai_process_map}

\begin{landscape}
\footnotesize
\setlength{\LTcapwidth}{\linewidth}
\begin{longtable}{@{}>{\raggedright\arraybackslash}p{0.098\linewidth}>{\raggedright\arraybackslash}p{0.140\linewidth}>{\raggedright\arraybackslash}p{0.140\linewidth}>{\raggedright\arraybackslash}p{0.098\linewidth}>{\raggedright\arraybackslash}p{0.082\linewidth}>{\centering\arraybackslash}p{0.025\linewidth}>{\centering\arraybackslash}p{0.025\linewidth}>{\centering\arraybackslash}p{0.025\linewidth}>{\centering\arraybackslash}p{0.066\linewidth}>{\centering\arraybackslash}p{0.072\linewidth}>{\centering\arraybackslash}p{0.052\linewidth}>{\centering\arraybackslash}p{0.052\linewidth}@{}}
\caption{CT suite process map for the cost of delivering care along the conventional CT pathway estimated using FL-TDABC}
\label{tab:ct_suite_with_ai}\\

\toprule
\textbf{Activity} & \textbf{Sub-tasks} & \textbf{Personnel} & \textbf{Equipment} & \textbf{Material} &
\textbf{a} & \textbf{b} & \textbf{c} & \textbf{Expected Time (min)} &
\textbf{Expected Cost (CAD)} & \textbf{Cost Low} & \textbf{Cost High} \\
\midrule
\endfirsthead

\toprule
\textbf{Activity} & \textbf{Sub-tasks} & \textbf{Personnel} & \textbf{Equipment} & \textbf{Material} &
\textbf{a} & \textbf{b} & \textbf{c} & \textbf{Expected Time (min)} &
\textbf{Expected Cost (CAD)} & \textbf{Cost Low} & \textbf{Cost High} \\
\midrule
\endhead

\bottomrule
\endlastfoot

\multicolumn{12}{c}{\cellcolor{gray!50}\textbf{CT Suite}} \\
\midrule

\multirow{3}{=}{Patient brought to CT} 
& Patient transported by ED nurse 
& ED nurse; CT nurse 
& Stretcher; Monitor 
& -- 
& 10 & 15 & 30 & 18.33 
& 32.44 & 17.7 & 53.1 \\
\cmidrule(lr){2-12}
& Monitored by: a) ED nurse, b) stroke team, c) CT tech 
&  &  &  &  &  &  &  &  &  & \\
\cmidrule(lr){2-12}
& d) ED physician e) ED tech 
&  &  &  &  &  &  &  &  &  & \\
\midrule

NCCT scan 
& NCCT done by CT tech 
& CT technologist; CT nurse 
& CTH (CT suite only); Monitor; Stretcher 
& -- 
& 4 & 6 & 8 & 6 
& 10.32 & 6.88 & 13.76 \\
\midrule

CT scan interpretation 
& CT scan read by radiologist 
& Radiologist; Stroke Neurologist 
& -- 
& -- 
& 5 & 7 & 15 & 9 
& 13.32 & 7.4 & 13.76 \\
\midrule

\multirow{2}{=}{Patient eligibility confirmed for IV-tPA} 
& tPA eligibility decision by a) stroke neurologist 
& Stroke Neurologist; Stroke Resident; ED physician; ED nurse 
& IV-tPA (Alteplase 100 mg) 
& -- 
& 5 & 10 & 30 & 15 
& 2809 & 2767 & 2872 \\
\cmidrule(lr){2-12}
& b) in house stroke team 
&  &  &  &  &  &  &  &  &  & \\
\midrule

Eligible, IV-tPA given / Not eligible 
& tPA mixed and administered 
&  &  &  &  &  &  &  &  &  & \\
\midrule

CTA scan 
& CT tech completes scan 
& CT technologist; CT nurse 
& CTH (CT suite only); Monitor; Stretcher 
& Contrast (CTA) 
& 2 & 4 & 10 & 5.33 
& 134.1676 & 128.44 & 142.2 \\
\midrule

CTA scan interpreted by AI
& -- 
& -- 
& --
& -- 
& -- & -- & -- & -- 
& -- & -- & -- \\
\midrule

\multirow{3}{=}{AI result verification} 
& AI result verified by radiologist 
& Radiologist 
& -- 
& -- 
& \multirow{3}{*}{3} & \multirow{5}{*}{8} & \multirow{3}{*}{15} & \multirow{3}{*}{9.33} 
& \multirow{3}{*}{13.80} & \multirow{3}{*}{7.4} & \multirow{3}{*}{22.2} \\
\cmidrule(lr){2-5}

& AI result verified by Neuro IR MD 
& Neuro IR MD 
& -- 
& -- 
&  &  &  &  &  &  & \\
\cmidrule(lr){2-5}

& AI result verified by Neurologist 
& Stroke Neurologist 
& -- 
& -- 
&  &  &  &  &  &  & \\
\midrule

LVO / No LVO 
& LVO confirmation 
& Radiologist; Stroke Neurologist 
& -- 
& -- 
& 5 & 8 & 15 & 9.33 
& 13.80 & 7.4 & 22.2 \\
\midrule

\end{longtable}
\end{landscape}

\end{document}